\documentclass[runningheads]{llncs}
\usepackage{graphicx}

\usepackage{tikz}
\usepackage{comment}
\usepackage{amsmath,amssymb} 
\usepackage{color}

\usepackage{bm} 
\usepackage{epsfig}
\usepackage{amsfonts}
\usepackage{url}
\usepackage{algorithm}
\usepackage{algorithmicx}
\usepackage{algpseudocode}
\usepackage{booktabs}
\usepackage{threeparttable}
\usepackage{multirow}
\usepackage{subcaption}
\usepackage{caption}
\usepackage{array}
\usepackage{soul}
\usepackage{footnote}
\usepackage{tablefootnote}
\usepackage{wrapfig}
\usepackage{bbding}

\DeclareMathOperator*{\argmax}{argmax}
\DeclareMathOperator*{\argmin}{argmin}

\usepackage[accsupp]{axessibility}  

\begin{document}
\pagestyle{headings}
\mainmatter
\def\ECCVSubNumber{4681}  

\title{ScaleNet: Searching for the Model to Scale\thanks{This work was supported in part by National Natural Science Foundation of China (NSFC) No. 61922015, U19B2036, and in part by Beijing Natural Science Foundation Project No. Z200002.}} 


\titlerunning{ScaleNet: Searching for the Model to Scale}
%
\author{
    Jiyang Xie\inst{1}\index{Xie, Jiyang} \and 
    Xiu Su\inst{2}\index{Su, Xiu} \and 
    Shan You\inst{3}\index{You, Shan} \and 
    Zhanyu Ma\inst{1}\index{Ma, Zhanyu} \and 
    Fei Wang\inst{4}\index{Wang, Fei} \and \\
    Chen Qian\inst{3}\index{Qian, Chen}
}
\authorrunning{J. Xie et al.}
%
\institute{
Pattern Recognition and Intelligent Systems Lab., Beijing University of Posts and Telecommunications, China. \email{\{xiejiyang2013,mazhanyu\}@bupt.edu.cn} \and
The University of Sydney, Australia. \email{xisu5992@uni.sydney.edu.au} \and
SenseTime Research Centre, China. \email{\{youshan,qianchen\}@sensetime.com} \and
University of Science and Technology of China. \email{wangfei91@mail.ustc.edu.cn}
}
\maketitle

\begin{abstract}
Recently, community has paid increasing attention on model scaling and contributed to developing a model family with a wide spectrum of scales. Current methods either simply resort to a one-shot NAS manner to construct a non-structural and non-scalable model family or rely on a manual yet fixed scaling strategy to scale an unnecessarily best base model. In this paper, we bridge both two components and propose ScaleNet to jointly search base model and scaling strategy so that the scaled large model can have more promising performance. Concretely, we design a super-supernet to embody models with different spectrum of sizes (\emph{e.g.}, FLOPs). Then, the scaling strategy can be learned interactively with the base model via a Markov chain-based evolution algorithm and generalized to develop even larger models. To obtain a decent super-supernet, we design a hierarchical sampling strategy to enhance its training sufficiency and alleviate the disturbance. Experimental results show our scaled networks enjoy significant performance superiority on various FLOPs, but with at least $2.53\times$ reduction on search cost. Codes are available at \url{https://github.com/luminolx/ScaleNet}.
\keywords{Neural architecture search (NAS), model scaling, hierarchical sampling strategy, Markov chain-based evolution algorithm}
\end{abstract}

\section{Introduction}\label{sec:intro}

\begin{figure}[!t]
    \begin{center}
        \includegraphics[width=.95\linewidth]{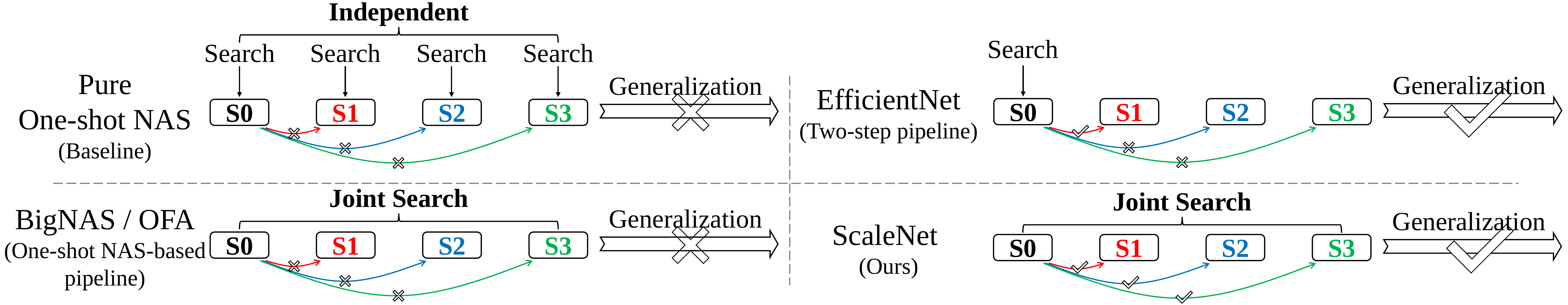}
    \end{center}
    \caption{Comparisons of different methods to generate a model family S$0$-S$3$. \textcolor{red}{Red}, \textcolor{blue}{blue}, and \textcolor{green}{green} arrows are the scaling strategy searching for scaling stage \textcolor{red}{$1$}, \textcolor{blue}{$2$}, and \textcolor{green}{$3$}, respectively. The pure one-shot NAS (baseline) independently searched various models without either scaling strategy modeling between scaling stages or larger-scale architecture generalization. BigNAS~\cite{yu2020bignas} jointly searched the model family from an entire supernet, although it obtained non-structural then non-scalable architectures. Then in EfficientNet~\cite{tan2019efficientnet}, the base model (\emph{i.e.}, S$0$) and the scaling strategy to S$1$ were only searched independently for architecture generalizing. In our ScaleNet, we combine the two components and jointly search base model and all scaling strategies to scale the model into infinite ones.}
    \label{fig:motivation}
\end{figure}

Convolutional neural networks (CNNs) have achieved great performance in computer vision with various model architectures~\cite{chang2021your,cheng2022sufficient,du2021progressive,he2016deep,howard2019searching,sandler2018mobilenetv2,xie2021gpca,xie2021advanced,xie2021ds-ui,xu2019automatic,you2017learning,zagoruyko2017wide} proposed for better feature extraction abilities. Previous work~\cite{su2021prioritized,su2021locally,su2021bcnet,su2021vitas,you2020greedynas} usually focused on how to automatically design a model architecture under a certain resource budget (\emph{e.g.}, floating-point operations per second, FLOPs) with neural architecture search (NAS) algorithms and gained significant improvements. However, due to different levels of budgets which may occur in various applications, multi-scale architectures should be considered in practice and can be independently generated by the NAS. Nevertheless, typical NAS methods~\cite{su2021prioritized,su2021locally,su2021bcnet,su2021vitas,you2020greedynas} have to search one a time for each scale, and the searching cost will be approximated linearly scaled as well~\cite{tan2021efficientnetv2} (see baseline in Figure~\ref{fig:motivation}).

In contrast, recent work~\cite{cai2020onceforall,dai2021fbnetv3,dollar2021fast,han2020model,li2021searching,liu2021greedy,lou2021dynamic,tan2019efficientnet,wan2020fbnetv2,wu2019fbnet,yu2020bignas,zhai2021scaling} get down to paying attention on model scale and designing a model family in a more straightforward way. Two frameworks have been proposed as shown in Figure~\ref{fig:motivation}, including one-shot NAS-based pipeline (\emph{e.g.}, BigNAS~\cite{yu2020bignas} and Once-for-All (OFA)~\cite{cai2020onceforall}) and two-step pipeline~(\emph{e.g.},~EfficientNet~\cite{tan2019efficientnet} and~EfficientNet-X~\cite{li2021searching}). The former directly designed an overcomplete one-shot supernet to embody multiple (finite) scales and searched models by NAS. However, they are difficult to extend the searched models to a larger one, since finding a specific scaling strategy that adapts all the non-structurally searched  architectures is infeasible. The latter decomposed the large model generation with two steps,~\emph{i.e.}, first acquiring an optimal base model, then scaling it on three dimensions, including depth, width, and resolution, using some pre-defined strategies,~\emph{e.g.}, compound scaling~\cite{tan2019efficientnet} and fast compound scaling~\cite{dollar2021fast}. However, the best base model is unnecessarily optimal for scaling. How to combine the advantages of both,~\emph{i.e.}, \textit{automatically and jointly searching the base model and scaling strategy by NAS and freely extending the scaling strategy into infinite scales}, should be carefully considered.

Different with manually designed rule-based scaling strategies~\cite{dollar2021fast,tan2019efficientnet,yang2022modeling}, we propose to \textit{directly discover} the optimal scaling strategies within base model search. One-shot NAS can search model architectures based on a trained supernet that contains all the possible architectures (so-called paths). For improving search efficiency, we apply an even-larger supernet dubbed super-supernet to embody multi-scale networks. However, a common one-shot space usually has a uni-modal distribution of FLOPs of paths under uniform sampling~\cite{guo2020single,yu2020bignas}. In this way, the super-supernet tends to favor the intermediate-FLOPs and cannot accommodate all FLOPs budgets well, which will in turn hampers the optimality of searched scaling strategies. Inspired by ancestral sampling~\cite{bishop2006pattern}, we propose a hierarchical sampling strategy (HSS) that splits the search space into partitions and the sampling is implemented respectively. The search space and the sampling distribution of the super-supernet for FLOPs are carefully designed according to the budgets of various scaling stages and undertake a multi-modal form distribution. 

Secondly, considering that our goal is to find a base model architecture with the strongest scaling capability (instead of the best performance) and its corresponding optimal scaling strategies, we propose a joint search for them, dubbed Markov chain-based evolution algorithm (MCEA), by iteratively and interactively optimizing both of them. After obtaining the searched scaling strategies, we model the trends of depth, width, and resolution, respectively, and generalize them to develop even larger models. We theoretically derive a group of generalization functions in the three dimensions for larger-scale architectures, with which moderate performance can be actually achieved.

The contributions of this paper are four-fold:
\begin{itemize}
    \item We propose ScaleNet to jointly search the base model and a group of the scaling strategies based on one-shot NAS framework. The scaling strategies of larger scales are generalized by the searched ones with our theoretically derived generalization functions.
    \item We carefully design the search space and a multi-modal distribution for FLOPs budgets for hierarchical sampling strategy (HSS) in the one-shot NAS-based scaling search algorithm to enhance the training sufficiency of paths in super-supernet. 
    \item We propose a joint search algorithm for both the base model and the scaling strategies, namely Markov chain-based evolution algorithm (MCEA), by iteratively and interactively optimizing both of them.
    \item Experimental results show that the searched architectures by the proposed ScaleNet with various FLOPs budgets can outperform the referred methods on various datasets including ImageNet-$1$k. Meanwhile, search time can be significantly reduced at least $2.53\times$. 
\end{itemize}

\begin{figure*}[!t]
    \begin{center}
        \includegraphics[width=.85\linewidth]{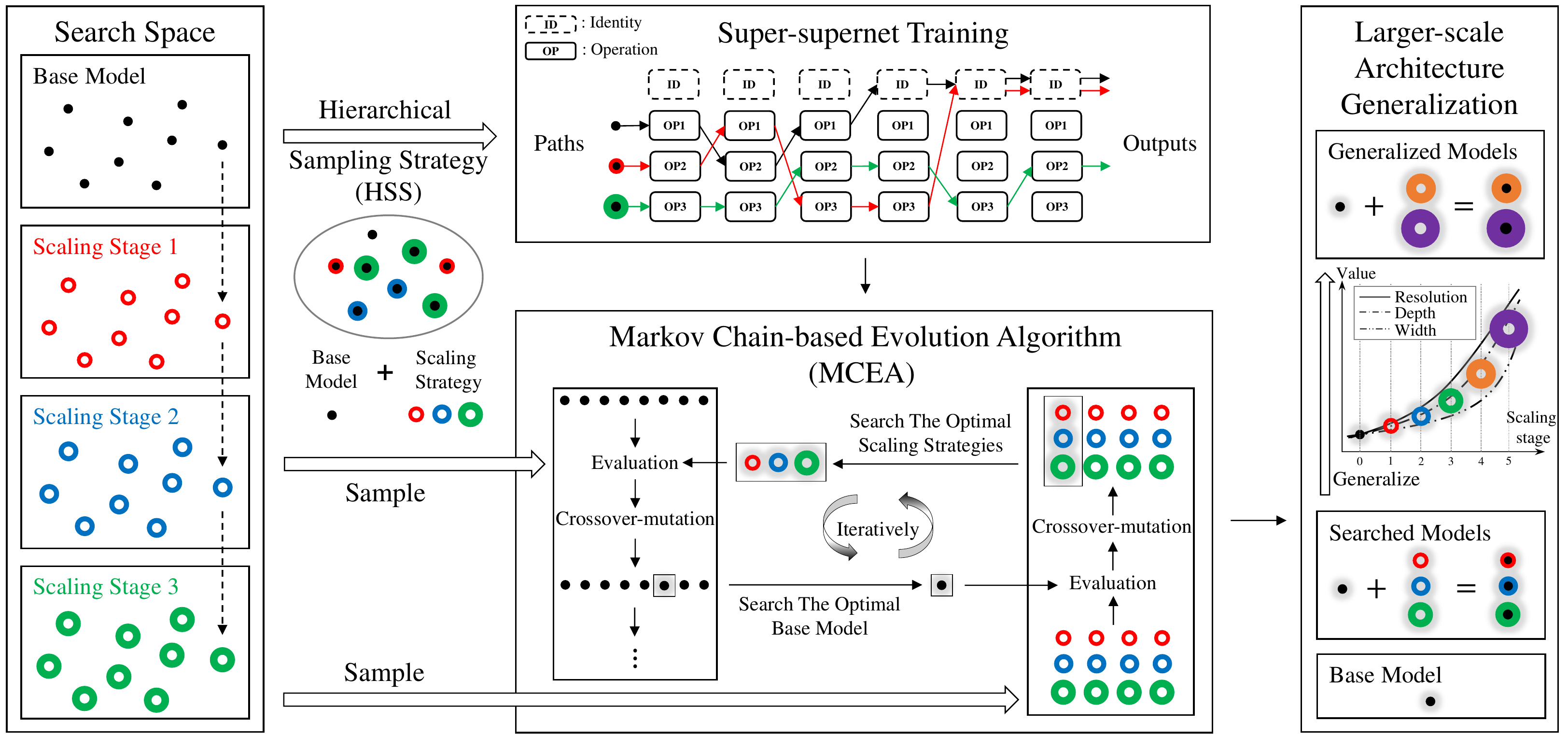}
    \end{center}
    \caption{Framework of the proposed ScaleNet. Based on the carefully designed search space with multiple scaling stage (the left box), we apply the proposed hierarchical sampling strategy (HSS) for sampling paths (one path is generated by a base model and a scaling strategy) in one-shot super-supernet training (the upper box). Then, we utilize the proposed Markov chain-based evolution algorithm (MCEA) to iteratively and iteractively search the optimal base model and scaling strategies (the lower box). In each iteration, an evolution procedure with crossover-mutation and evaluation is undertaken for searching the optimal base model or scaling strategies based on the search space. Finally, after obtaining the optimal ones, we generalize them to larger-scale architectures by the estimations of the trends of depth, width, and resolution, respectively (the right box). All the obtained architectures will be applied for retraining and inference.}
    \label{fig:method}
\end{figure*}

\section{Related Work}\label{sec:related_work}
\subsection{One-shot NAS-based Model Family Searching}\label{ssec:nas}

FBNets~\cite{dai2021fbnetv3,wan2020fbnetv2,wu2019fbnet} optimized CNN architectures for mobile devices and generated a family of models in order to avoid training individual architectures separately and reduce resource consuming. Cai~\emph{et al}.~\cite{cai2020onceforall} proposed to train a once-for-all (OFA) model that supports diverse architectural settings by decoupling training and search, in order to reduce the cost. BigNAS~\cite{yu2020bignas} challenged the conventional pipeline that post-processing of model weights is necessary to achieve good performance and constructed big single-stage models without extra retraining or post-processing. 
However, the main drawback of these methods is that they only searched for a model family by training a joint or even a group of independent supernet(s), but did not analyze the structural relationship and explicit scaling strategies between the architectures with different budgets in the model family. It is difficult and even infeasible to extend the scaling strategies to larger scales.

\subsection{Model Scaling}\label{ssec:scaling}



Tan and Le~\cite{tan2019efficientnet} systematically studied model scaling and found that carefully balancing depth, width, and resolution of a model can lead to better performance. They proposed to empirically obtain the optimal compound scaling that effectively scales a specific base model up to gain a model family,~\emph{i.e.}, EfficientNet. Its variant versions, such as EfficientNetV$2$~\cite{tan2021efficientnetv2} and EfficientNet-X~\cite{li2021searching}, improved it in trade-off on speed and accuracy. A simple fast compound scaling strategy~\cite{dollar2021fast} was proposed to encourage to primarily scale model width, while scaling depth and resolution to a lesser extent for memory efficiency. Another work~\cite{liu2021greedy} built an greedy network enlarging method based on the reallocation of computations in order to enlarge the capacity of CNNs by improving the three dimensions on stage level. 
However, the aforementioned work always estimated the optimal scaling strategy for theoretical double-FLOPs budget by a small grid search, which is computationally expensive and does not match the actual FLOPs budgets. Meanwhile, they only considered to find the strategy for the smallest scaling stage, which did not learn the dependency between larger scaling stages. Furthermore, the relation between a base model and scaling strategies did not be investigated, which means the base is not the optimal for scaling.

\section{ScaleNet}\label{sec:scalenet}

Due to the drawbacks of the one-shot NAS-based model family searching and compound scaling-based model scaling, we propose to combine their advantages together and fill the gap between them. Here, ScaleNet jointly searches a base model with the strongest scaling capability and the optimal scaling strategies based on one-shot NAS framework by training a super-supernet as shown in Figure~\ref{fig:method}. The super-supernet training and joint searching procedures are carefully designed for the goal. Then, when obtaining the searched scaling strategies, we model the trends of depth, width, and resolution, respectively, and generalize to develop even larger architectures. All the searched and generalized scaling strategies will be applied for the final model family construction and training.

\subsection{One-shot Joint Search Space for Model Scaling}\label{ssec:search_space}

\begin{wrapfigure}{r}{.5\linewidth}
    \centering
    \includegraphics[width=.95\linewidth]{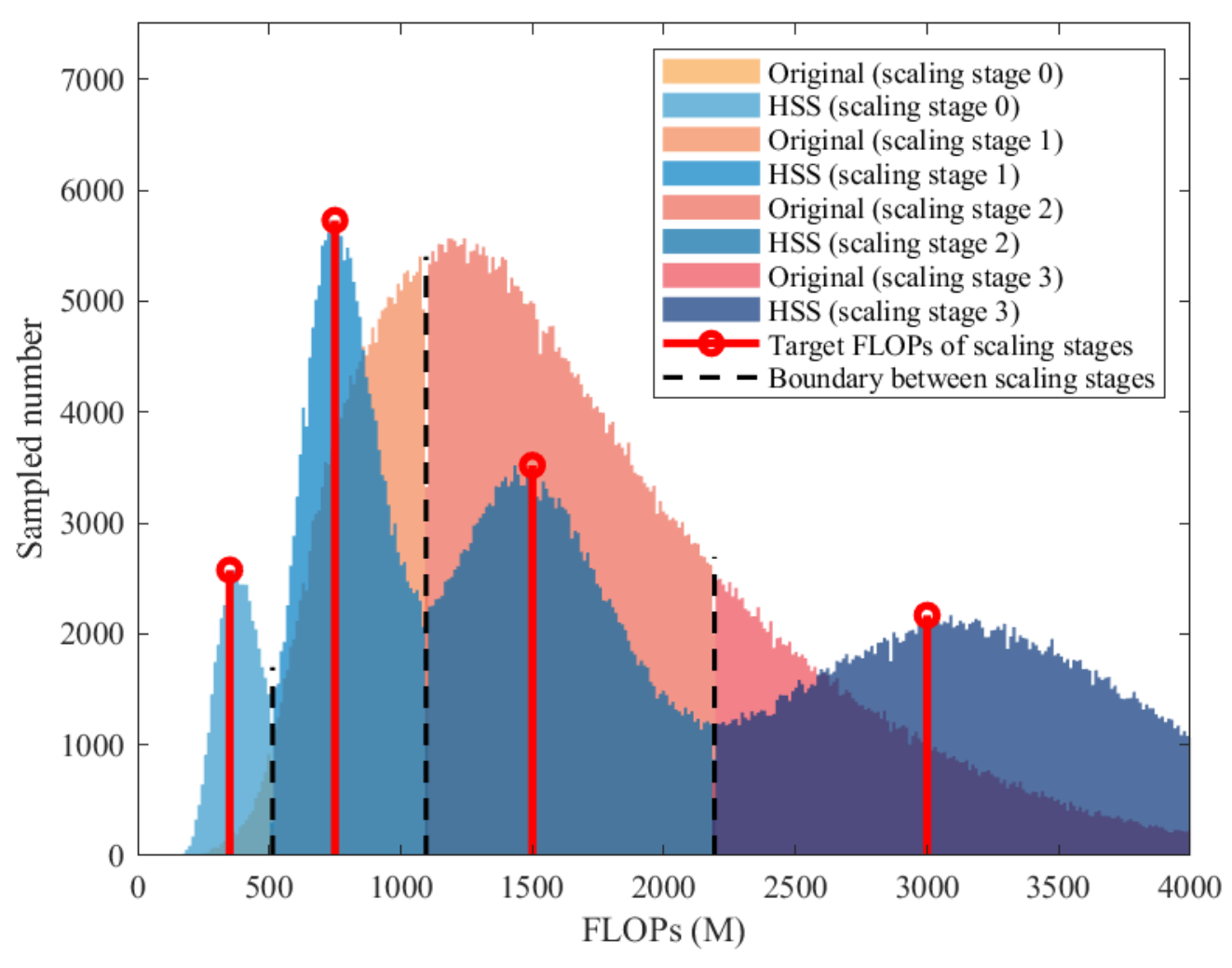}
    \caption{Sampling distribution based on the proposed HSS, compared with that of the original uniform sampling in~\cite{yu2020bignas}. We took $750,000$ paths for each to simulate actual super-supernet training.}
    \label{fig:search_space}
\end{wrapfigure}

The FLOPs budget of the base model is selected according to the mean FLOPs of the search space as shown in Figure~\ref{fig:search_space} (Detailed information of the search space are shown in supplementary material). Then for various scaling stages, their FLOPs budgets are exponentially expanded by that of the base model $\alpha$. As scaling strategies $\left\{\boldsymbol{S}_j=[d_j,w_j,r_j]\right\}_{j=0}^M$ with maximum scaling stage as $M$, including the change ratios of depth $d_j$, width $w_j$, and resolution $r_j$ (real numbers that are not smaller than one), have their corresponding FLOPs budgets, respectively, we assign a scaling strategy to each base model architecture to compute the mean FLOPs and find the center point of the search space of scaling stage $j$ according to its FLOPs budget. The detailed settings of the whole search space are elaborated in the supplementary material.

\subsection{Hierarchical Sampling Strategy for Super-supernet Training}\label{ssec:sampling}

Applying the original one-shot NAS framework~\cite{yu2020bignas,guo2020single} by utilizing uniform sampling to train the super-supernet means each operation has equal probability to be selected. It has two disadvantages: $1$) the original bell-shape sampling distribution towards one specific FLOPs budget for the whole search space is not suitable for the multiple scaling stages with various budgets, where paths cannot be fairly trained in each scaling stage (see the red histogram in Figure~\ref{fig:search_space}); $2$) the search space is almost $300\times$ larger and the size of super-supernet is $8\times$ larger (statistics with base model and three scaling stages trained) than before, which increase the difficulty of super-supernet training. As we need to search architectures in different scaling stages, the paths with the less selected FLOPs budgets in super-supernet training are not sufficiently trained and those in one scaling stage are not fairly trained. Here, we propose a hierarchical sampling strategy (HSS) by implementing a multi-modal sampling distribution to address the above issues (see the blue histogram in Figure~\ref{fig:search_space}). 

As we assigned $d_j$, $w_j$, and $r_j$ into scaling stages in the search space, we treat the target multi-modal sampling distribution $p(\boldsymbol{\alpha}, \boldsymbol{S})$ as a mixture model,~\emph{i.e.},
\begin{equation}
    p(\boldsymbol{\alpha}, \boldsymbol{S})=p(\boldsymbol{\alpha})\cdot p(\boldsymbol{S})=\begin{matrix}p(\boldsymbol{\alpha})\cdot\left(\sum_{j=0}^M\eta_j p_j(\boldsymbol{S})\right)\end{matrix},
\end{equation}

\noindent where $p(\boldsymbol{\alpha})$ and $p_j(\boldsymbol{S})$ are the sampling distributions of base model $\boldsymbol{\alpha}$ and scaling stage $j$, respectively, and $\eta_j$ is normalized component weight, $\sum_{j=0}^M\eta_j=1$ and $\eta_j\ge0$. Here, we can empirically set equal component weights,~\emph{i.e.}, $\eta_j=\tfrac{1}{M+1}$, or normalized combination ratios of scaling strategies in the search space. 

In sampling, we apply the ancestral sampling of a probabilistic graphic model~\cite{bishop2006pattern} that is a two-step hierarchical strategy for the scaling strategies. We firstly select a scaling stage $m$ given the conditional distribution $p(m|\eta_1,\cdots,\eta_M)$, which is a categorical distribution as
\begin{equation}
    p(m=j|\eta_1,\cdots,\eta_M)=\eta_j.
\end{equation}

\noindent Then, we uniformly sample a scaling strategy $\boldsymbol{S}$ in scaling stage $m$. Meanwhile, a base model $\boldsymbol{\alpha}$ is sampled as well based on the original uniform sampling.

The one-shot super-supernet training process based on the proposed HSS is
\begin{equation}
    \boldsymbol{W}^{*}=\argmin_{\boldsymbol{W}}loss_{\boldsymbol{\alpha},\boldsymbol{S}\sim p(\boldsymbol{\alpha},\boldsymbol{S})}\left(\boldsymbol{W}(\boldsymbol{\alpha},\boldsymbol{S});\boldsymbol{D}_{train}\right),
\end{equation}

\noindent where $\boldsymbol{W}$ is a set of super-supernet parameters, $\boldsymbol{W}^{*}$ is a set of the optimal parameters, $loss$ is training loss function (commonly cross-entropy loss), $\boldsymbol{W}(\boldsymbol{\alpha},\boldsymbol{S})$ means a path that is constructed by $\boldsymbol{\alpha}$ and $\boldsymbol{S}$, and $\boldsymbol{D}_{train}$ is training set.

\subsection{Interactive~Search~for~Base~Model~and~Multiple~Scaling~Strategy}\label{ssec:search}

After completing the super-supernet training, we usually search both base model and a group of scaling strategies by an evolution algorithm (EA). The original objective of the optimization is globally maximizing the weighted sum of the validation accuracies $ACC$ of the $M$ scaling stages~\cite{wang2019evolutionary}] as
\begin{align}\label{eq:jointobj}
    &\begin{matrix}\max_{\boldsymbol{\alpha}, \{\boldsymbol{S}_j\}}\left[\sum_{j=1}^M \pi_j ACC\left(\boldsymbol{W}^{*}(\boldsymbol{\alpha},\boldsymbol{S}_j);\boldsymbol{D}_{val}\right)\right]\end{matrix},\\
    \text{s.t.}&\;\;\boldsymbol{\alpha}\in\boldsymbol{\Omega}_{\boldsymbol{\alpha}},\boldsymbol{S}_j\in\boldsymbol{\Omega}_j,\text{FLOPs}(\boldsymbol{W}^{*}(\boldsymbol{\alpha},\boldsymbol{S}_j))=f_j,\nonumber
\end{align}

\noindent where $\pi_j$ is the normalized weight of scaling stage $j$ (constrained by $\sum_{j=1}^M\pi_j=1$), $ACC$ is the validation accuracy of a path, $\boldsymbol{D}_{val}$ is validation set, $\boldsymbol{\Omega}_{\boldsymbol{\alpha}}$ and $\boldsymbol{\Omega}_j$ are the search space of base model and scaling stage $j$, respectively, and $f_j$ is the FLOPs budget of scaling stage $j$. 

However, both the too large search space and constrained computational resource restrict the search. Meanwhile, searching the globally optimal group of architectures from the search space is difficult and expensive, since redundant information and noises may affect the search. 

\begin{figure}[!t]
    \centering
    \includegraphics[width=.85\linewidth]{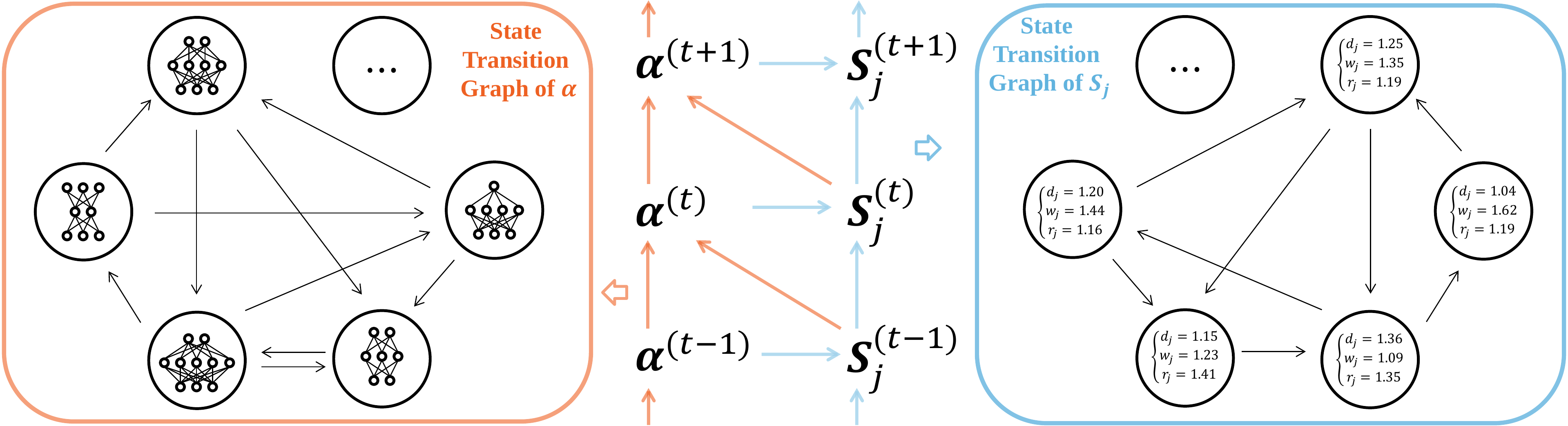}
    \caption{The interactive search in the proposed Markov chain-based evolution algorithm (MCEA) with a coupled Markov chain. The coupled Markov chain combines both the left (\eqref{eq:astar}-\eqref{eq:amc}) and the right (\eqref{eq:sstar}-\eqref{eq:smc}) ones for the interaction.}
    \label{fig:mcea}
\end{figure}

Here, inspired by Markov process, we propose a so-called Markov chain-based evolution algorithm (MCEA) with a coupled Markov chain, which iteratively and interactively optimizes $\boldsymbol{\alpha}$ and $\{\boldsymbol{S}_j\}$, to overcome the global search issue. As shown in Figure~\ref{fig:mcea}, with maximum iteration number $T$ and iteration index $t=1,\cdots,T$, we transfer the optimization problem in~\eqref{eq:jointobj} to iteratively and interactively solving the limiting distributions $\gamma(\boldsymbol{\alpha})$ and $\gamma(\{\boldsymbol{S}_j\})$ to obtain the optimal $\boldsymbol{\alpha}^*$ and $\boldsymbol{S}_j^*$, respectively, as 
\begin{align}
    \boldsymbol{\alpha}^*
    =&\argmax_{\boldsymbol{\alpha}}\gamma(\boldsymbol{\alpha})
    =\begin{matrix}\argmax_{\boldsymbol{\alpha}}\left[\lim_{t\to\infty}p(\boldsymbol{\alpha}^{(t)})\right]\end{matrix},\label{eq:astar}\\
    p(\boldsymbol{\alpha}^{(t)})
    =&\begin{matrix}\sum_{\boldsymbol{\alpha}^{(t-1)}}p(\boldsymbol{\alpha}^{(t)}|\boldsymbol{\alpha}^{(t-1)},\{\boldsymbol{S}_j\}_{j=1}^M=\{\boldsymbol{S}_j^{(t-1)}\}_{j=1}^M)p(\boldsymbol{\alpha}^{(t-1)})\end{matrix},\label{eq:amc}\\
    \boldsymbol{S}_j^*
    =&\argmax_{\boldsymbol{S}_j}\gamma(\boldsymbol{S}_j)
    =\begin{matrix}\argmax_{\boldsymbol{S}_j}\left[\lim_{t\to\infty}p(\boldsymbol{S}_j^{(t)})\right]\end{matrix},\label{eq:sstar}\\
    p(\boldsymbol{S}_j^{(t)})
    =&\begin{matrix}\sum_{\boldsymbol{S}_j^{(t-1)}}p(\boldsymbol{S}_j^{(t)}|\boldsymbol{S}_j^{(t-1)},\boldsymbol{\alpha}=\boldsymbol{\alpha}^{(t)})p(\boldsymbol{S}_j^{(t-1)})\end{matrix},\label{eq:smc}
\end{align}

\noindent where $p(\boldsymbol{\alpha}^{(t-1)})$ and $p(\boldsymbol{S}_j^{(t-1)})$ are the state probabilities of the discrete variable $\boldsymbol{\alpha}^{(t-1)}$ and $\boldsymbol{S}_j^{(t-1)}$, respectively, with the enumerable search space as state space. $p(\boldsymbol{\alpha}^{(t)}|\boldsymbol{\alpha}^{(t-1)},\{\boldsymbol{S}_j\}_{j=1}^M=\{\boldsymbol{S}_j^{(t-1)}\}_{j=1}^M)\propto\sum_{j=1}^M \pi_j ACC\left(\boldsymbol{W}^{*}(\boldsymbol{\alpha},\boldsymbol{S}_j^{(t-1)});\boldsymbol{D}_{val}\right)$ and $p(\boldsymbol{S}_j^{(t)}|\boldsymbol{S}_j^{(t)},\boldsymbol{\alpha}=\boldsymbol{\alpha}^{(t-1)})\propto ACC\left(\boldsymbol{W}^{*}(\boldsymbol{\alpha}^{(t)},\boldsymbol{S}_j);\boldsymbol{D}_{val}\right)$ are transition matrices and are approximated implemented by the crossover-mutation process under FLOPs budgets given the obtained scaling strategies or base model, respectively. $\boldsymbol{S}_j^{(0)}$ is the initial scaling strategy of the $j^{th}$ scaling stage, where probability 
$p(\boldsymbol{S}_j^{(0)})\propto \tfrac{1}{K}\sum_{k=1}^K ACC\left(\boldsymbol{W}^{*}(\boldsymbol{\alpha}_k^{(0)},\boldsymbol{S}_j);\boldsymbol{D}_{val}\right)$ is obtained by a group of randomly selected base models $\{\boldsymbol{\alpha}_k^{(0)}\}_{k=1}^K$ under the FLOPs budget of base model.

\subsection{Larger-scale Architecture Generalization with Searched Scaling Strategies}\label{ssec:prediction}

The scaling strategies of larger scales are generalized by the searched ones. We define the optimal scales of $M+1$ scaling stages as $\{\boldsymbol{\hat{S}}_j\}_{j=0}^M$. We should note that we pre-define $\boldsymbol{\hat{S}}_0$ as $\hat{d}_0=\hat{w}_0=\hat{r}_0=1$ for the base model. 

We argue that depth, width, and resolution should have distinct growth rates, respectively, since they perform different roles in the model scaling. Meanwhile, $j$ is exponentially proportional to FLOPs budgets in our setting, but $\boldsymbol{\hat{S}}_j,j=1,\cdots,M$ are under almost linear- or quadratic-level. Therefore, inspired by~\cite{tan2019efficientnet}, we propose to utilize the independent regression functions for depth, width, and resolution for the larger-scale generalization with
\begin{equation}
    \left\{
        \begin{array}{l}
            \hat{d}_j=a_0^{(d)}\cdot\left(\left(a_1^{(d)}\right)^j-1\right)+1\\
            \hat{w}_j=a_0^{(w)}\cdot\left(\left(a_1^{(w)}\right)^j-1\right)+1\\
            \hat{r}_j=a_0^{(r)}\cdot\left(\left(a_1^{(r)}\right)^j-1\right)+1
        \end{array}
    \right.\label{eq:dwr_final}
\end{equation}

\noindent to guarantee the values in $\boldsymbol{\hat{S}}_0$, where $a_0$ and $a_1$ are parameters which can be directly optimized by stochastic gradient descent (SGD) or other optimization algorithms. As we can learn different values of the parameters, respectively, the three dimensions can obtain distinct growth rates. 

\textbf{Derivation of Larger-scale Architecture Generalization Functions.} We define FLOPs budget as $f$ here. We can obtain the relation between $f$ and scaling stage $j$ as $f\propto\ 2^{\theta\times j}$ where $\propto$ means ``proportional to'', $\theta>0$ is a parameter. As the depth $\hat{d}$, width $\hat{w}$, resolution $\hat{r}$ are positively correlated with $f$, we have $\theta^{(d)}$, $\theta^{(w)}$, and $\theta^{(r)}$ with $\theta=\theta^{(d)}+\theta^{(w)}+\theta^{(r)}$ for the three, and obtain
\begin{equation}
    2^{(\theta^{(d)}+\theta^{(w)}+\theta^{(r)})\times j}\ \dot{\propto}\ \hat{d}\times\hat{w}^2\times\hat{r}^2,
\end{equation}

\noindent where $\dot{\propto}$ means ``positively correlated with'', but not ``proportional to''. We formulate the relation between $j$ and $\hat{d}$ as $2^{\theta^{(d)}\times j}\ \dot{\propto}\ \hat{d}$ and introduce a linear approximation as 
$2^{\theta^{(d)}\times j}\approx\beta \hat{d}+\delta \Rightarrow
    \hat{d}\approx\tfrac{1}{\beta}\cdot 2^{\theta^{(d)}\times j}-\tfrac{\delta}{\beta}$, 
where $\beta$ and $\delta$ are parameters. Here, we define 
$a_0^{(d)}=\tfrac{1}{\beta},\ a_1^{(d)}=2^{\theta^{(d)}},\ a_2^{(d)}=-\tfrac{\delta}{\beta}$ 
and obtain
\begin{equation}
    \hat{d}=a_0^{(d)}\cdot \left(a_1^{(d)}\right)^{j}+a_2^{(d)}.\label{eq:d}
\end{equation}

\noindent Note that as $a_0^{(d)}$, $a_1^{(d)}$, and $a_2^{(d)}$ are parameters, ``$\approx$'' can be transferred to ``$=$''.

\setlength{\tabcolsep}{12pt}
\begin{table}[!t]
  \centering
  \caption{Comparisons with other state-of-the-art methods on ImageNet-$1$k dataset. Top-$1$ and Top-$5$ accuracies (\%), FLOPs (G), and numbers of parameters (\#Param., M) are reported. 
  The best results are highlighted in \textbf{bold}.
  }
    \begin{tabular}{lcccc}
    \hline
    \multicolumn{1}{c}{Model} & Top-$1$ & Top-$5$ & FLOPs & \#Param. \\
    \hline
    FBNetV$2$-L$1$~\cite{wan2020fbnetv2} & $77.2$ & N/A & $0.33$ & N/A \\
    OFA-$80$~\cite{cai2020onceforall} & $76.8$ & $93.3$ & $0.35$ & $6.1$ \\
    GreedyNAS-A~\cite{you2020greedynas} & $77.1$ & $93.3$ & $0.37$ & $6.5$ \\
    EfficientNet-B$0$~\cite{tan2019efficientnet} & $76.3$ & $93.2$ & $0.39$ & $5.3$\\
    \bf ScaleNet-S$\boldsymbol{0}$ (ours) & $\boldsymbol{77.5}$ & $\boldsymbol{93.7}$ & $\boldsymbol{0.35}$ & $\boldsymbol{4.4}$ \\
    \hline
    EfficientNet-B$1$~\cite{tan2019efficientnet} & $78.8$ & $94.4$ & $0.70$ & $7.8$ \\
    OFA-$200$~\cite{cai2020onceforall} & $79.0$ & $94.5$ & $0.78$ & $11.0$ \\
    RegNetY-$800$MF~\cite{radosavovic2020designing} & $76.3$ & N/A & $0.80$ & $6.3$ \\
    EfficientNet-X-B$0$~\cite{li2021searching} & $77.3$ & N/A & $0.91$ & $7.6$ \\
    \bf ScaleNet-S$\boldsymbol{1}$ (ours) & $\boldsymbol{79.9}$ & $\boldsymbol{94.8}$ & $\boldsymbol{0.80}$ & $\boldsymbol{7.4}$ \\
    \hline
    EfficientNet-B$2$~\cite{tan2019efficientnet} & $79.8$ & $94.9$ & $1.00$ & $9.2$ \\
    EfficientNet-B$2$ (re-impl) & $80.4$ & $95.1$ & $1.00$ & $9.2$ \\
    BigNASModel-XL~\cite{yu2020bignas} & $80.9$ & N/A & $1.04$ & $9.5$ \\
    EfficientNet-X-B$1$~\cite{li2021searching} & $79.4$ & N/A & $1.58$ & $9.6$ \\
    \bf ScaleNet-S$\boldsymbol{2}$ (ours) & $\boldsymbol{81.3}$ & $\boldsymbol{95.6}$ & $\boldsymbol{1.45}$ & $\boldsymbol{10.2}$ \\
    \hline
    EfficientNet-B$3$~\cite{tan2019efficientnet} & $81.1$ & $95.5$ & $1.80$ & $12.0$ \\
    EfficientNet-X-B$2$~\cite{li2021searching} & $80.0$ & N/A & $2.30$ & $10.0$ \\
    RegNetY-$3.2$GF~\cite{radosavovic2020designing} & $79.0$ & N/A & $3.20$ & $19.4$ \\
    RegNetY-$4$GF~\cite{radosavovic2020designing} & $79.4$ & N/A & $4.00$ & $20.6$ \\
    \bf ScaleNet-S$\boldsymbol{3}$ (ours) & $\boldsymbol{82.2}$ & $\boldsymbol{95.9}$ & $\boldsymbol{2.76}$ & $\boldsymbol{13.2}$ \\
    \hline
    RegNetY-$500$M$\rightarrow\!4$GF~\cite{dollar2021fast} & $81.7$ & N/A & $4.10$ & $36.2$ \\
    EfficientNet-B$4$~\cite{tan2019efficientnet} & $82.6$ & $96.3$ & $4.20$ & $19.0$ \\
    EfficientNet-X-B$3$~\cite{li2021searching} & $81.4$ & N/A & $4.30$ & $13.3$ \\
    RegNetY-$8$GF~\cite{radosavovic2020designing} & $81.7$ & N/A & $8.00$ & $39.2$ \\
    \bf ScaleNet-S$\boldsymbol{4}$ (ours) & $\boldsymbol{83.2}$ & $\boldsymbol{96.6}$ & $\boldsymbol{5.97}$ & $\boldsymbol{16.1}$ \\
    \hline
    EfficientNet-B$5$~\cite{tan2019efficientnet} & $83.3$ & $96.7$ & $9.90$ & $30.0$ \\
    EfficientNet-X-B$4$~\cite{li2021searching} & $83.0$ & N/A & $10.40$ & $21.6$ \\
    RegNetY-$500$M$\rightarrow\!16$GF~\cite{dollar2021fast} & $83.1$ & N/A & $16.20$ & $134.8$ \\
    EfficientNet-B$0\rightarrow\!16$GF~\cite{dollar2021fast} & $83.2$ & N/A & $16.20$ & $122.8$ \\
    \bf ScaleNet-S$\boldsymbol{5}$ (ours) & $\boldsymbol{83.7}$ & $\boldsymbol{97.1}$ & $\boldsymbol{10.22}$ & $\boldsymbol{20.9}$ \\
    \hline
    \end{tabular}
  \label{tab:sota}
\end{table}
\setlength{\tabcolsep}{1.4pt}

Then, due to $\hat{d}=1$ for the base model (\emph{i.e.}, scaling stage $0$), we should guarantee the relation. Thus, we put $\hat{d}=1,j=0$ into~\eqref{eq:d} and obtain 
$a_2^{(d)}=1-a_0^{(d)}$. 
We re-put it into~\eqref{eq:d} and obtain the depth function in~\eqref{eq:dwr_final} as
\begin{align}
    \hat{d}
    =a_0^{(d)}\cdot \left(a_1^{(d)}\right)^{j}+\left(1-a_0^{(d)}\right)
    =a_0^{(d)}\cdot\left(\left(a_1^{(d)}\right)^j-1\right)+1.
\end{align}

Similarly, we can achieve the relation between $i$ and $\hat{w}$, $\hat{r}$, respectively, as
\begin{align}
    \hat{w}&\approx\sqrt{\tfrac{1}{\beta'}\cdot 2^{\theta^{(w)}\times i}-\tfrac{\delta'}{\beta'}}\approx\tfrac{1}{\sqrt{\beta'}}\sqrt{2}^{\theta^{(w)}\times i}-\sqrt{\tfrac{\delta'}{\beta'}},\\
    \hat{r}&\approx\sqrt{\tfrac{1}{\beta''}\cdot 2^{\theta^{(r)}\times i}-\tfrac{\delta''}{\beta''}}\approx\tfrac{1}{\sqrt{\beta''}}\sqrt{2}^{\theta^{(r)}\times i}-\sqrt{\tfrac{\delta''}{\beta''}},
\end{align}

\noindent where $\beta'$, $\delta'$, $\beta''$, and $\delta''$ are parameters. 

\section{Experimental Results and Discussions}\label{sec:experiments}

\subsection{Performance of ScaleNet on ImageNet-$\boldsymbol{1}$k}\label{ssec:imagenet}

We conducted experiments on ImageNet-$1$k dataset~\cite{russakovsky2015imagenet} for the proposed ScaleNet with recently proposed methods. Note that we divided a mini validation set ($50$ images per class) from the training set for evaluation in the MCEA. The search models are named by S$j$, where S$0$ is the base model, S$1$, S$2$, and S$3$ are searched by the MCEA, and S$4$ and S$5$ are the generalized ones. The FLOPs budgets are selected according to Figure~\ref{fig:search_space}. Detailed settings are in the supplementary material. In Table~\ref{tab:sota} with different FLOPs budgets, the searched models of our ScaleNet can achieve the best performance among those with similar FLOPs.

\setlength{\tabcolsep}{18pt}
\begin{table}[!t]
  \centering
  \caption{Performance in five fine-tuning tasks. Top-$1$ accuracies (\%), FLOPs (G), parameter numbers (\#Param., M) are reported. The best results are in \textbf{bold}.}
    \begin{tabular}{l@{}l@{}c@{}c@{}c}
    \hline
    \multicolumn{1}{c}{Dataset} & \multicolumn{1}{c}{Model} &\ \ Top-$1$\ \ &\ \ FLOPs\ \ &\ \ \#Param.\ \ \\
    \hline
    \multirow{3}[0]{*}{FGVC Aircraft~\cite{maji2013finegrained}\ \ } 
    & Inception-v$4$~\cite{szegedy2017inceptionv4} & $90.9$ & $13.00$ & $41$ \\
    & EfficientNet-B$3$~\cite{tan2019efficientnet} & $90.7$ & $1.80$ & $10$ \\
    & \bf ScaleNet-S$\boldsymbol{3}$ (ours) & $\boldsymbol{91.4}$ & $\boldsymbol{2.76}$ & $\boldsymbol{11}$ \\
    \hline
    \multirow{3}[0]{*}{Stanford Cars~\cite{krause20133d}} 
    & Inception-v$4$~\cite{szegedy2017inceptionv4} & $93.4$ & $13.00$ & $41$ \\
    & EfficientNet-B$3$~\cite{tan2019efficientnet} & $93.6$ & $1.80$ & $10$ \\
    & \bf ScaleNet-S$\boldsymbol{3}$ (ours) & $\boldsymbol{94.4}$ & $\boldsymbol{2.76}$ & $\boldsymbol{11}$ \\
    \hline
    \multirow{3}[0]{*}{Food-$101$~\cite{bossard2014food101}} 
    & Inception-v$4$~\cite{szegedy2017inceptionv4} & $90.8$ & $13.00$ & $41$ \\
    & EfficientNet-B$4$~\cite{tan2019efficientnet} & $91.5$ & $4.20$ & $17$ \\
    & \bf ScaleNet-S$\boldsymbol{4}$ (ours) & $\boldsymbol{92.0}$ & $\boldsymbol{5.97}$ & $\boldsymbol{14}$ \\
    \hline
    \multirow{3}[0]{*}{CIFAR-$10$~\cite{krizhevsky09cifar}} 
    & NASNet-A~\cite{zoph2018learning} & $98.0$ & $42.00$ & $85$ \\
    & EfficientNet-B$0$~\cite{tan2019efficientnet} & $98.1$ & $0.39$ & $4$ \\
    & \bf ScaleNet-S$\boldsymbol{0}$ (ours) & $\boldsymbol{98.3}$ & $\boldsymbol{0.35}$ & $\boldsymbol{3}$ \\
    \hline
    \multirow{3}[0]{*}{CIFAR-$100$~\cite{krizhevsky09cifar}} 
    & NASNet-A~\cite{zoph2018learning} & $87.5$ & $42.00$ & $85$ \\
    & EfficientNet-B$0$~\cite{tan2019efficientnet} & $88.1$ & $0.39$ & $4$ \\
    & \bf ScaleNet-S$\boldsymbol{0}$ (ours) & $\boldsymbol{88.4}$ & $\boldsymbol{0.35}$ & $\boldsymbol{3}$ \\
    \hline
    \end{tabular}
  \label{tab:downstream}
\end{table}
\setlength{\tabcolsep}{1.4pt}

\setlength{\tabcolsep}{10pt}
\begin{table}[!t]
  \centering
  \caption{Ablation studies. Top-$1$ accuracies (\%) of s$0$-s$4$ models on ImageNet-$100$ dataset are reported. In column ``Sampl.'', ``U'' and ``H'' are the original uniform sampling and the proposed HSS, respectively. $M$ and $T$ are the maximum scaling stage and the iteration of the proposed MCEA. ``Val'' (for $T$ only) means the validation accuracy (\%) in the MCEA. The best results are shown in \textbf{bold}.}
    \begin{tabular}{@{}c@{}c@{}c@{}c@{}c@{}c@{}c@{}c@{}c@{}}
    \hline
    \ Sampl.\ \ &\ \ $M$\ \ &\ \ $T$\ \ &\ \ \ \ \ Val\ \ \ \ \ &\ \ \ \ \ s$0$\ \ \ \ \ &\ \ \ \ \ s$1$\ \ \ \ \ &\ \ \ \ \ s$2$\ \ \ \ \ &\ \ \ \ \ s$3$\ \ \ \ \ &\ \ \ \ \ s$4$\ \ \ \ \\
    \hline
    U & $3$ & $4$ & N/A & $84.16$ & $86.96$ & $87.72$ & $89.26$ & $90.02$ \\
    \hline
    H & $1$ & $4$ & N/A & $84.06$ & $86.30$ & $87.93$ & $88.86$ & $90.30$ \\
    H & $2$ & $4$ & N/A & $84.42$ & $86.24$ & $88.02$ & $89.12$ & $90.14$ \\
    \hline
    H & $3$ & $1$ & $63.58$ & $84.18$ & $86.34$ & $\boldsymbol{88.12}$ & $88.90$ & $89.76$ \\
    H & $3$ & $2$ & $63.38$ & $84.20$ & $85.86$ & $88.00$ & $89.44$ & $90.18$ \\
    H & $3$ & $4$ & $\boldsymbol{63.61}$ & $\boldsymbol{84.76}$ & $\boldsymbol{87.18}$ & $88.10$ & $\boldsymbol{89.90}$ & $90.46$ \\
    H & $3$ & $6$ & $63.59$ & $84.44$ & $86.42$ & $87.80$ & $89.54$ & $\boldsymbol{90.48}$ \\
    H & $3$ & $8$ & $63.53$ & $84.50$ & $86.48$ & $87.64$ & $89.30$ & $90.36$ \\
    \hline
    \end{tabular}
  \label{tab:ablation}
\end{table}
\setlength{\tabcolsep}{1.4pt}

\subsection{Transferability to Fine-tuning Tasks}\label{ssec:transfer_learning}

In addition to the experiments on ImageNet-$1$k, we also transferred the searched architectures to fine-tuning tasks 
by fine-tuning our ImageNet-pretrained models. Experimental settings can be found in the supplementary material. Table~\ref{tab:downstream} shows the transfer learning results. Ours can outperform different referred models, respectively. When applying larger models, we can gain further improvement.

\begin{wrapfigure}{r}{.5\linewidth}
    \centering
    \includegraphics[width=.8\linewidth]{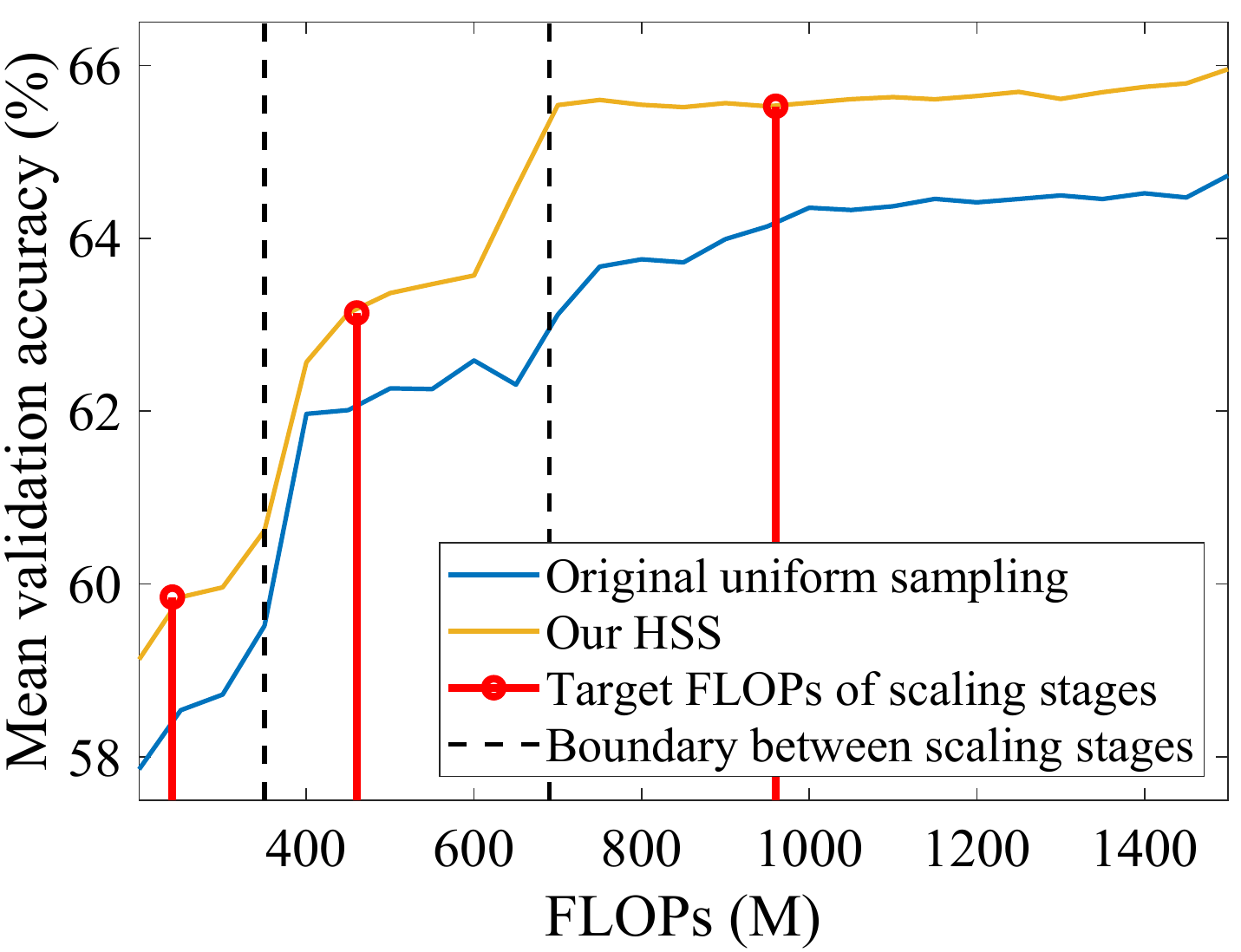}
    \caption{Mean validation accuracies in the $4^{th}$ iteration of the MCEA by using original uniform sampling in~\cite{yu2020bignas} and our HSS, respectively. The accuracies are grouped based on FLOPs. Each non-overlapping group contains recent $50$M FLOPs paths. We merely show the performance of the three scaling stages, as we only evaluate them in the MCEA, except those of the base model.}
    \label{fig:hss_val_acc}
\end{wrapfigure}

\subsection{Ablation Studies}\label{ssec:ablation}

We discuss the effect of the proposed components for the ScaleNet on ImageNet-$100$ dataset~\cite{russakovsky2015imagenet,tian2020contrastive}. We divided a mini validation set ($50$ images per class) from the training set. All the following validation accuracies were calculated by the mini one. Searched models are named by s$j$, where s$0$ ($120$M FLOPs) is the base model, s$1$ ($240$M FLOPs), s$2$ ($480$M FLOPs), and s$3$ ($960$M FLOPs) are searched by the MCEA, and s$4$ ($1920$M FLOPs) is the generalized one. The experimental results are shown in Table~\ref{tab:ablation}. Detailed experimental settings and visualization of search space are in supplementary material.

\textbf{Effect of HSS:} We compare the proposed HSS with the original uniform sampling in~\cite{yu2020bignas}. Our HSS improves the searched results with better retrained accuracies of s$0$-s$4$.

Furthermore, we illustrate the validation accuracies of paths by using both of them to evaluate the sufficiency of the super-supernet training. In Figure~\ref{fig:hss_val_acc}, the accuracies of our HSS are generally larger than those of the original one in addition to the FLOPs interval of $[360,440]$, as the interval is the mode of the original uniform sampling distribution. This means the proposed HSS can improve the sufficiency of the super-supernet training. 

We further analyze the Pearson, Spearman, and Kendall coefficients for validation accuracies of the two ones, respectively. All of them are the larger the better. Detailed settings are in the supplementary material. Table~\ref{tab:hss_metric} shows the values of our HSS significantly outperform the corresponding ones of the original. Specifically, our Pearson one is more than double of the original's.

\setlength{\tabcolsep}{10pt}
\begin{table}[!t]
  \centering
  \caption{Performance comparison of three coefficients (\%), including Pearson, Spearman, and Kendall coefficients, for validation accuracies by using original uniform sampling (``Original'') in~\cite{yu2020bignas} and our HSS, respectively, to evaluate the sampling strategies in super-supernet training. We sampled $6,000$ paths.}
    \begin{tabular}{lccc}
    \hline
    \multicolumn{1}{c}{Method} & Pearson & Spearman & Kendall \\
    \hline
    Original & $35.3$ & $80.1$ & $64.1$ \\
    \bf Our HSS & $\boldsymbol{73.6}$ & $\boldsymbol{83.9}$ & $\boldsymbol{66.2}$ \\
    \hline
    \end{tabular}
  \label{tab:hss_metric}
\end{table}
\setlength{\tabcolsep}{1.4pt}

\begin{figure}[!t]
    \centering
    \begin{subfigure}[t]{.22\linewidth}
        \centering
        \includegraphics[width=1\linewidth]{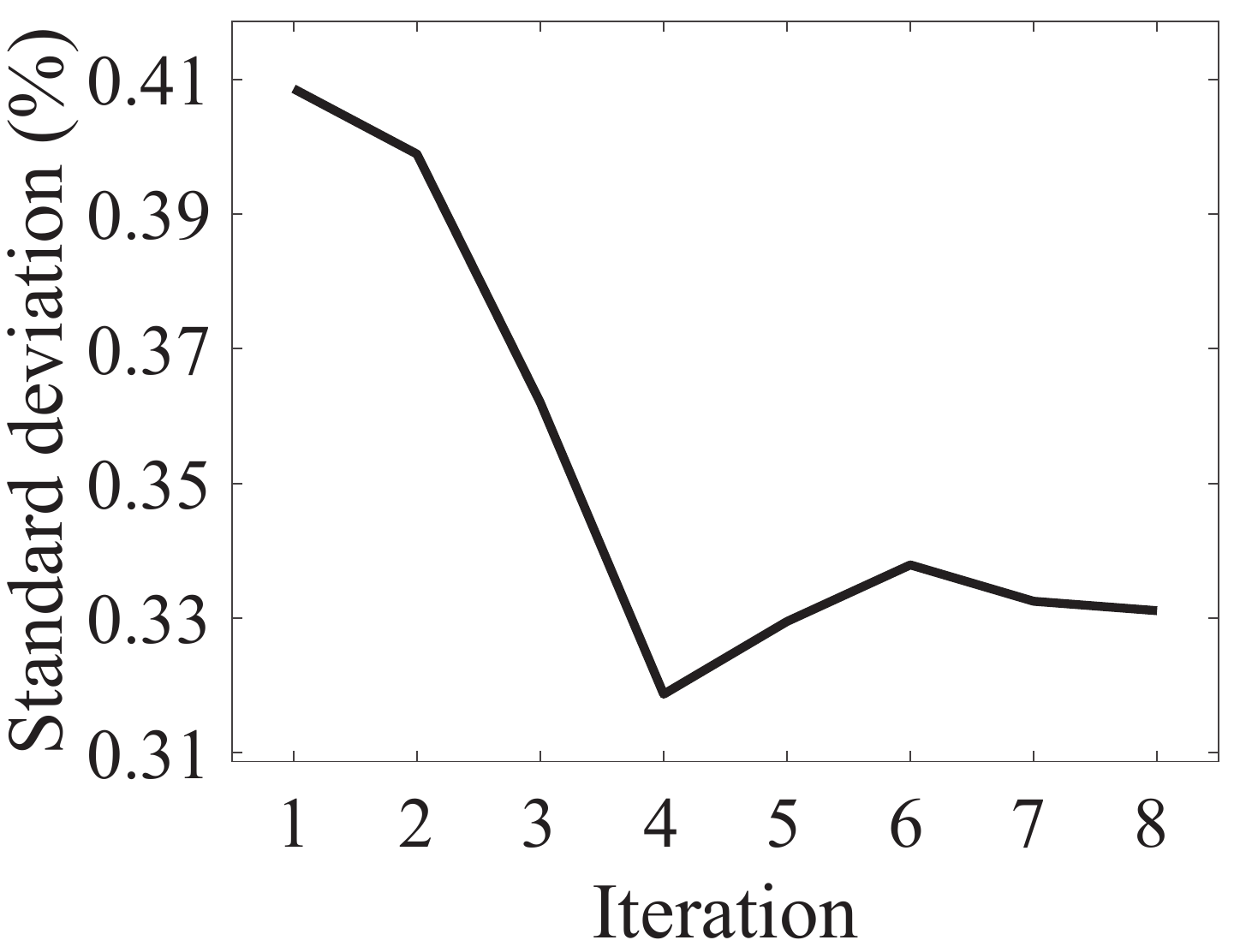}
        \subcaption{Base model.}
    \end{subfigure}
    \begin{subfigure}[t]{.22\linewidth}
        \centering
        \includegraphics[width=1\linewidth]{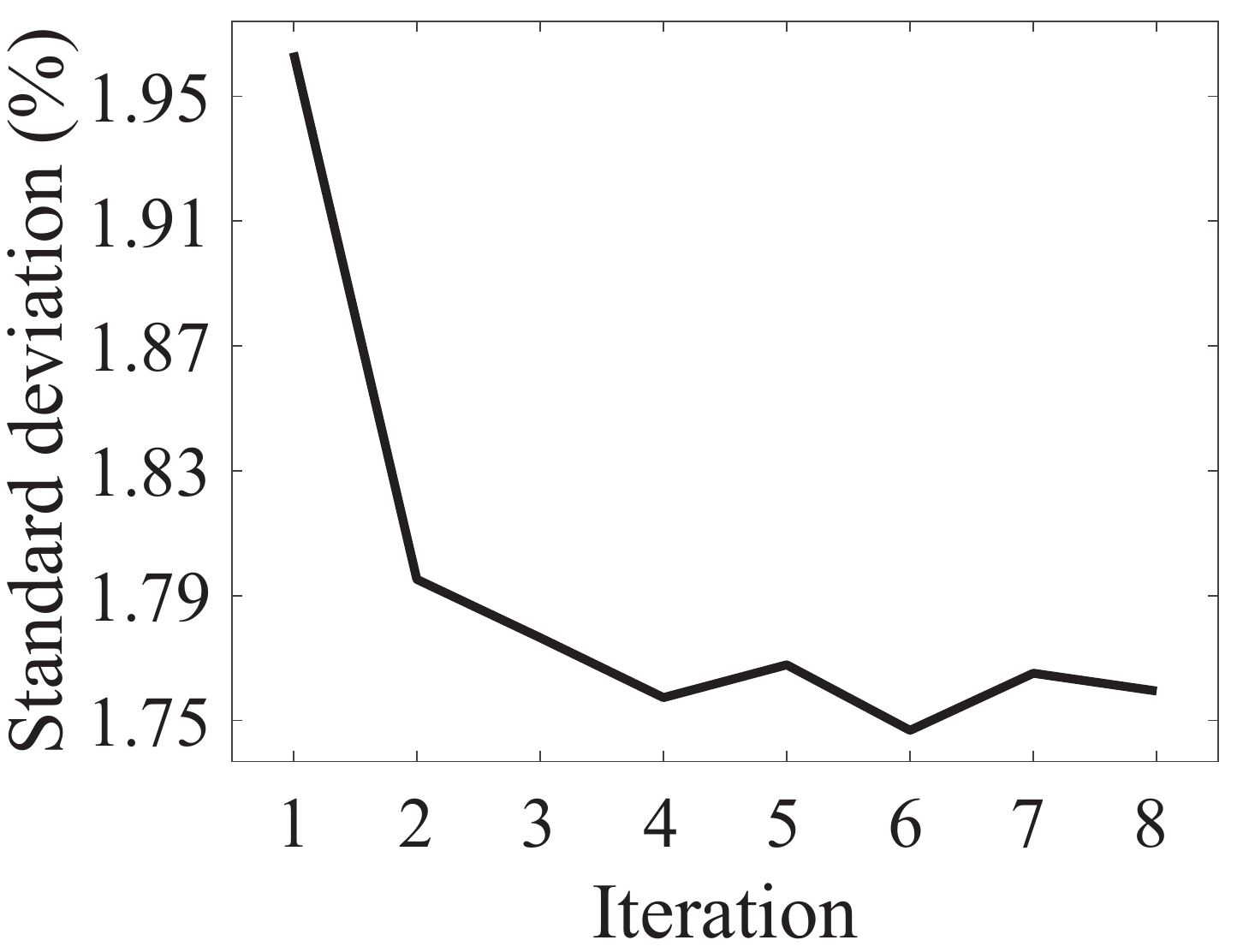}
        \subcaption{Scaling stage $1$.}
    \end{subfigure}
    \begin{subfigure}[t]{.215\linewidth}
        \centering
        \includegraphics[width=1\linewidth]{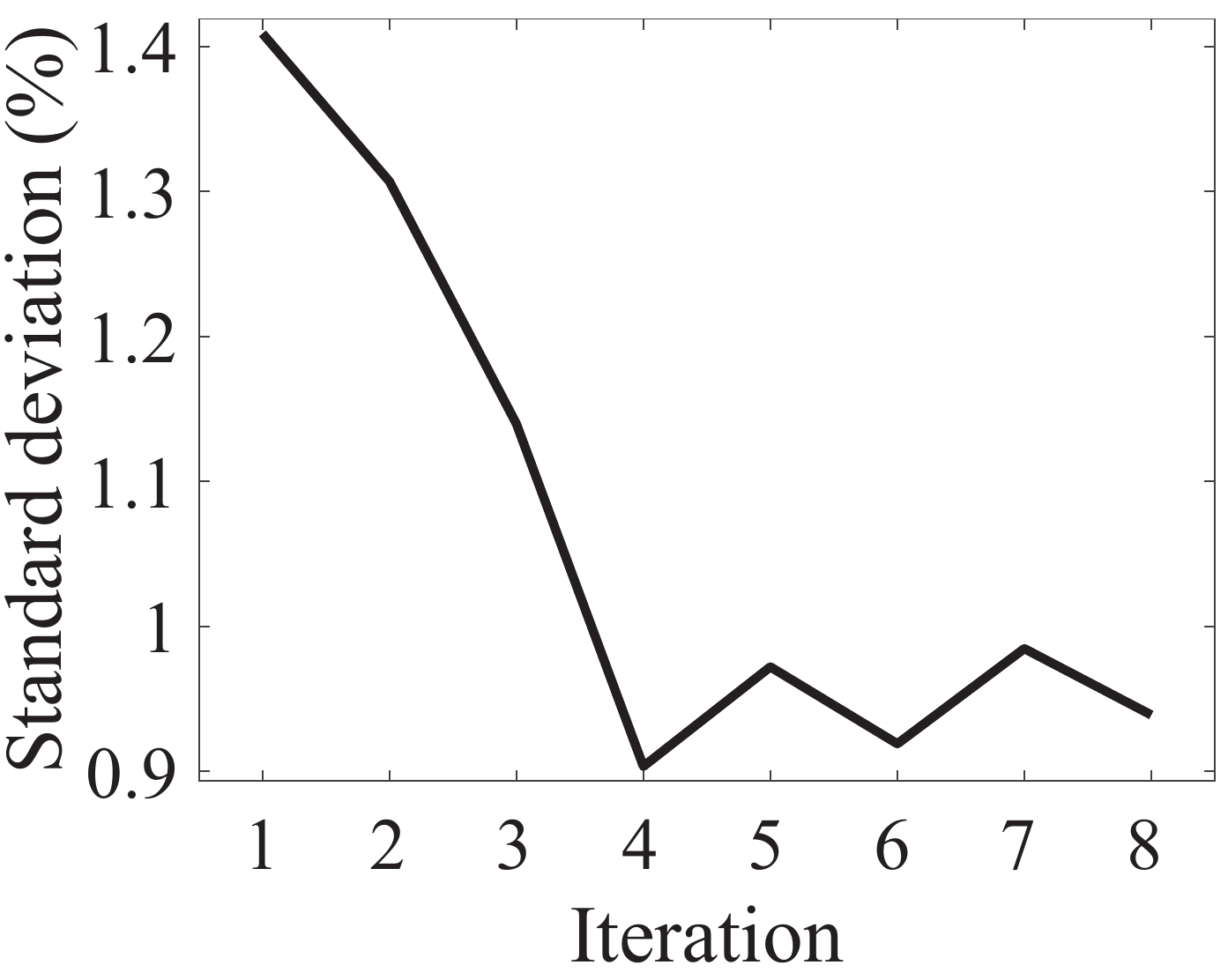}
        \subcaption{Scaling stage $2$.}
    \end{subfigure}
    \begin{subfigure}[t]{.22\linewidth}
        \centering
        \includegraphics[width=1\linewidth]{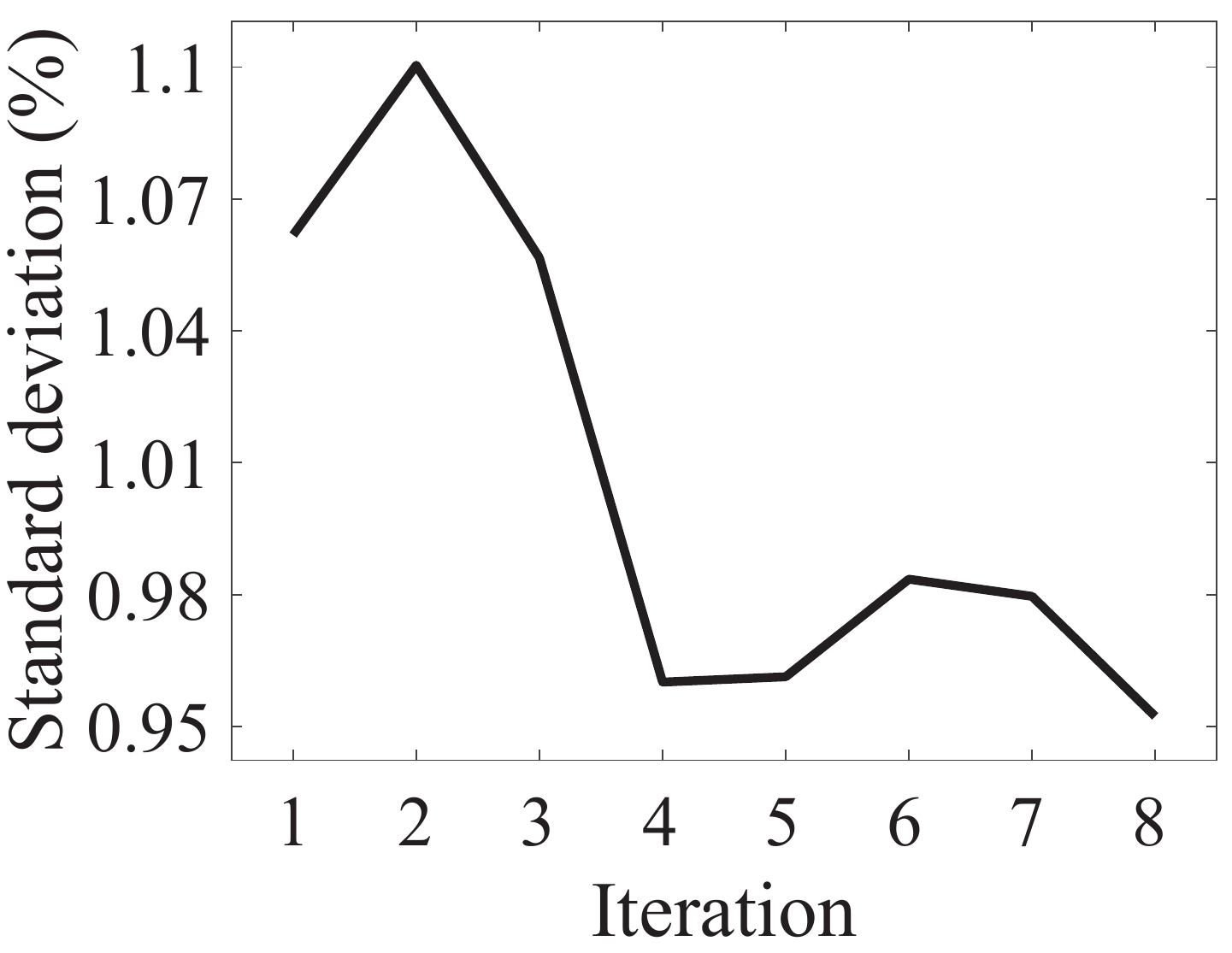}
        \subcaption{Scaling stage $3$.}
    \end{subfigure}
    \caption{Standard deviations (Stds) of validation accuracies in each iteration of the MCEA. In different scaling stages, the Stds decrease significantly in the first four iterations, while they then tend to be steady and convergent.}
    \label{fig:iter}
\end{figure}

\textbf{Effect of Maximum Scaling Stage in Searching:} We set the maximum scaling stage $M$ to be one, two, or three in the MCEA for searching. ScaleNet can gain better performance with larger $M$, which means a more suitable base model for scaling can be found. More scaling stages can achieve better performance by obtaining better base model architecture for scaling, which is a common sense. This means we do not have to validate with much larger $M$.

\textbf{Effect of Iteration in Searching:} We set the iteration $T$ as one, two, four, six, or eight in the MCEA for searching. When increasing $T$ from one to four, better base models and scaling strategies can be obtained with top-$1$ accuracies improved in most of the scaling stages. This means that larger $T$ can improve the searched results. However, when increasing $T$ from four to eight, similar performance can be found. This means about four iterations is enough for the search. Meanwhile, we can find that the retraining accuracies of s$0$-$4$ is relative to the validation accuracy in the MCEA, which shows the effectiveness of it. 

In addition, we show the standard deviations (Stds) of validation accuracies in each iteration of the MCEA in Figure~\ref{fig:iter} to analyze their convergence. In different scaling stages, the Stds decrease significantly in the first four iterations, while they then tend to be steady and convergent. This shows that our ScaleNet can effectively search the optimal ones and gradually minimize the Stds. 

\begin{figure}[!t]
    \centering
    \includegraphics[width=.95\linewidth]{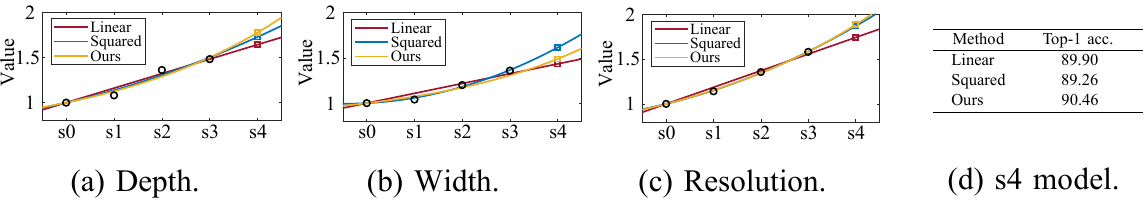}
    \caption{Comparison of larger-scale architecture generalization functions. The definitions of the compared ones are in the supplementary material. We generalized the three ones of s$4$ and retrained the scaled models on ImageNet-$100$ in (d).}
    \label{fig:comparison_dwr_prediction}
\end{figure}

\textbf{Effect of Larger-scale Architecture Generalization:} We experimentally compare the proposed exponential one with commonly used polynomial functions, such as linear and squared ones. As shown in Figure~\ref{fig:comparison_dwr_prediction}(a)-(c), three cases can precisely fit the trends of the depth, width, and resolution, respectively. Ours function can perform with different trends in the three dimensions. For depth and resolution, ours obtains rapid increase similar to the squared one, while it achieves gradual changes for width as the linear one. The total trends of ours are various, which is similar to the conclusion in~\cite{tan2019efficientnet}, but the other two always perform the uppermost or the lowest, which are unreasonable. 

We also trained all the scaled s$4$ models with three generalized scaling strategies, respectively, shown in Figure~\ref{fig:comparison_dwr_prediction}(d). The proposed one can achieve the best top-$1$ performance as $90.46\%$, superior to the other two functions. This shows the effectiveness of our larger-scale architecture generalization.

\setlength{\tabcolsep}{4pt}
\begin{table*}[!t]
  \centering
  \caption{Comparisons of search cost (GPU/TPU days). Those of supernet (super-supernet) training and searching are compared, respectively. ``Ratio-to-ScaleNet'' is the ratio between total cost of a model to that of the proposed ScaleNet, the smaller the better. ``$\dagger$'' means we estimated the lower-bound time. ``N/A'' means the work does not have the step. ``*'' means the work has the step but did not specifically mentioned in the paper. The best result is in \textbf{bold}.}
  \footnotesize
    \begin{tabular}{lcccccc}
    \hline
    \multicolumn{1}{c}{Model} & Device & Training & Searching & Total & Ratio-to-ScaleNet \\
    \hline
    MnasNet~\cite{tan2019mnasnet}$^{\dagger}$ & TPUv$2$ & N/A & $211,571$ & $211,571$ & $436.23\times$ & \\
    EfficientNet-X~\cite{li2021searching}$^{\dagger}$ & TPUv$3$ & N/A & $>1,765$ & $>1,765$ & $>3.64\times$ \\
    EfficientNet~\cite{tan2019efficientnet}$^{\dagger}$ & TPUv$3$ & N/A & $>1,714$ & $>1,714$ & $>3.53\times$ \\
    FBNetV$2$~\cite{wan2020fbnetv2}$^{\dagger}$ & V$100$ & * & * & $>1,633$ & $>3.37\times$ \\
    OFA~\cite{cai2020onceforall}$^{\dagger}$ & V$100$ & * & * & $>1,486$ & $>3.06\times$ \\
    BigNAS~\cite{yu2020bignas}$^{\dagger}$ & TPUv$3$ & $>960$ & $>268$ & $>1,228$ & $>2.53\times$ \\
    \bf ScaleNet (ours) & \bf V$\boldsymbol{100}$ & \bf $\boldsymbol{379}$ & \bf $\boldsymbol{106}$ & \bf $\boldsymbol{485}$ & \bf $\boldsymbol{1\times}$ \\
    \hline
    \end{tabular}
  \label{tab:search_cost}
\end{table*}
\setlength{\tabcolsep}{1.4pt}

\subsection{Discussion of Search Cost}\label{ssec:search_cost}

We discuss the efficiency of our ScaleNet, compared with a few recent strategies, including both one-shot NAS-based and two-step pipelines. We estimated the search cost for the referred ones under our FLOPs budgets, as they applied with various FLOPs budgets. The estimations are all shown in the supplementary material. As shown in Table~\ref{tab:search_cost}, the proposed ScaleNet can remarkably reduce the total search cost, which contains the cost of (super-)supernet training and searching. It can decrease at least $2.53\times$ and even $436.23\times$. Meanwhile, the proposed ScaleNet in the two parts of cost can still significantly improve the efficiency, respectively. Note that we used V$100$ for our experiments, while some others utilized TPUv$3$, which are much better than ours. This means we can achieve a larger decrease on the total search time under same resource conditions.


\begin{wrapfigure}{r}{.5\linewidth}
    \centering
    \includegraphics[width=.8\linewidth]{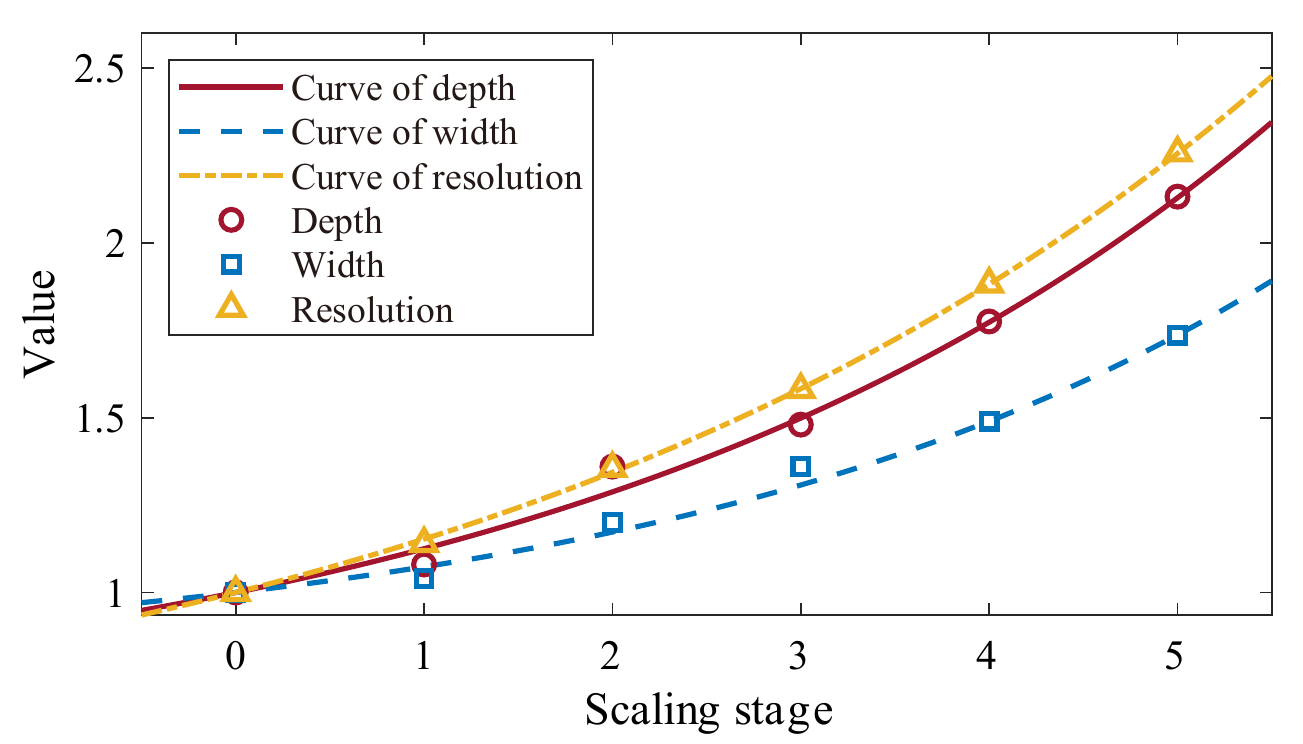}
    \caption{Trends of depth, width, and resolution on ImageNet-$1$k.}\label{fig:trend_dwr}
\end{wrapfigure}

\subsection{Discussion of the Trend of Scaling Strategies}\label{ssec:dwr_trend}

We discuss the trend of learned scaling strategies for different scaling stages (Figure~\ref{fig:trend_dwr}) in order to promote further scaling strategy design. 

\begin{itemize}
    \item Depth, width, and resolution change in an exponential order. They have different values of their change rates among scaling stages. Depth and resolution change similarly, while that of width is slightly smaller, which is similar to EfficientNet~\cite{tan2019efficientnet}.
\end{itemize}

\begin{itemize}
    \item Their values are not completely restricted by the theoretical constraint in~\cite{tan2019efficientnet}, but merely focus on the actual FLOPs budgets in each scaling stage. This means the searching process of ours is more fair.
\end{itemize}

\begin{itemize}
    \item After obtaining the searched ones that are good enough under the corresponding FLOPs budgets, the extended ones can be precisely constructed in FLOPs of generalized architectures as well under our estimation and work well in experiments.
\end{itemize}

\section{Conclusion}\label{sec:conclusion}

In this paper, we proposed ScaleNet to jointly search the base model and a group of the scaling strategies based on one-shot NAS framework. We improved the super-supernet training by the proposed HSS. Then, we jointly searched the base model and the scaling strategies by the proposed MCEA. The scaling strategies of larger scales were decently generalized by the searched ones. Experimental results show that the searched architectures by the proposed ScaleNet with various FLOPs budgets can outperform the referred methods on various datasets, including ImageNet-$1$k and fine-tuning tasks. Meanwhile, the searching time can be significantly reduced, compared with those one-shot NAS-based and manually designed two-step pipelines. 




\appendix

\section{Detailed Experimental Settings}\label{sec:settings}

In this section, we describe the detailed experimental settings of the proposed ScaleNet for super-supernet training, jointly base model and scaling strategies searching, model retraining, and fine-tuning tasks, respectively.

\textbf{FLOPs Budgets:} For both ImageNet-$1$k and ImageNet-$100$ datasets~\cite{russakovsky2015imagenet}, we used same method to determine the FLOPs budgets for various scaling stages. 

We first selected the FLOPs budget of base model by Monte Carlo simulation according to the search space in Table~\ref{tab:search_space_base}. We randomly sampled about $100,000$ paths of base model and calculated the mean FLOPs of them. We chose a multiple of $50$M near the mean as the FLOPs budget $f_0$ of base model. For scaling stage $j$, we selected $2^{j}\cdot f_0$ as the FLOPs budget and searched the optimal scaling strategy near it.

\textbf{Super-supernet Training:} For ImageNet-$1$k, we sampled the base model from the search space in Table~\ref{tab:search_space_base} and the scaling strategies in Table~\ref{tab:search_space_scaling}. We used a stochastic gradient descent (SGD) optimizer with momentum as $0.9$ and weight decay as $4\times10^{-5}$ to train the super-supernet. Initial learning was set as $0.12$ with a cosine annealing strategy for $750,000$ iterations. Learning rate warm-up was also included for $3,750$ iterations, linearly increasing from zero to $0.2$. We train the super-supernet with batch size as $1024$. In data augmentation, we randomly resized the batches according to the resolution values in the sampled scaling strategies, with common hyper-parameters of the augmentation. Then, we resized the batch to the scaled resolutions again. The maximum scaling stage $M$ was set as $3$ in the experiments.

\begin{figure}[!t]
    \begin{center}
        \includegraphics[width=.85\linewidth]{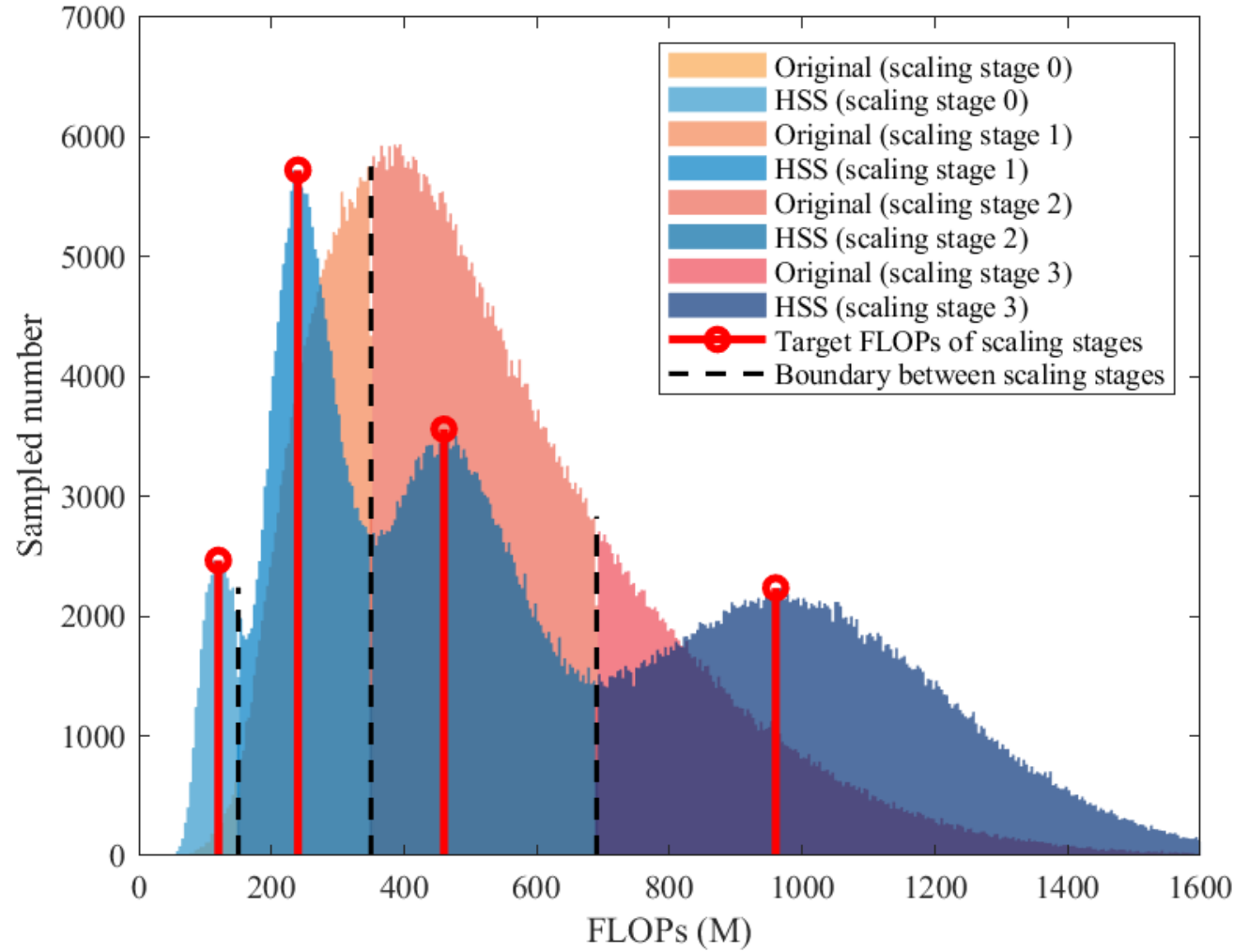}
    \end{center}
    \caption{Sampling distribution based on the proposed HSS, compared with that of the original uniform sampling for ablation studies (ImageNet-$100$). We took $750,000$ paths for each to simulate actual super-supernet training. The sampling distribution of the original uniform sampling performs as a bell-shape single-modal one, where paths cannot be fairly trained In each scaling stage (between black dashed lines). Meanwhile, paths in scaling stage $0$ and $3$ cannot be sufficiently trained due to the too small sampling probabilities of them. In contrary, our proposed HSS provides single-modal distributions for each scaling stage and the modes of each locate near the corresponding target FLOPs budgets.}
    \label{fig:search_space_ablation}
\end{figure}

For ImageNet-$100$ dataset in ablation studies, we followed the class selection~\cite{tian2020contrastive} and reduced the channel number of the super-supernet to a half with batch size as $256$, total iterations as $150,000$ and warm-up for $375$ iterations. All the other settings are same to those for ImageNet-$1$k experiments. The visualization of search space in the ablation studies is shown in Figure~\ref{fig:search_space_ablation}, which has similar trends with that for ImageNet-$1$k experiments.

We divided a mini validation set ($50$ images per class) from the training set for evaluation. The rest images in the training set were all used for super-supernet training. All the validation accuracies that we have emphasized in the paper were calculated by the mini validation set, respectively.

\begin{table*}[!t]
  \centering
  \caption{The macro-structure of the search space of the base model. ``$n$'' is the number of stacked building blocks, where ``$n_{\text{min}}$'' and ``$n_{\text{max}}$'' are the minimum and maximum numbers, respectively. ``Operation'' is the type of the block. ``N/A'' means ``not applied''. ``Y/N'' means ``using or not'' in the base model of the super-supernet training. ``Input'' is the original resolution of input feature maps. ``Channel'' is the number of output channels. ``Stride'' is the stride of the first block. ``Scale'' means which dimensions of the stage need to be scaled, where ``D'' is depth, ``W\_I'' is input channel number, ``W\_O'' is output channel number, ``R'' is resolution, ``\Checkmark'' means having the part, ``\XSolidBrush'' means ignoring the one. ``FC'' is a fully connected layer.}
    \resizebox{1\linewidth}{!}{
    \begin{tabular}{c|c|c|c|c|c|c|c|c|c|c|c|c}
    \hline
    \multirow{2}[0]{*}{Stage} & \multicolumn{2}{c|}{$n$} & \multirow{2}[0]{*}{Operation} & \multirow{2}[0]{*}{Expand rate} & \multirow{2}[0]{*}{SE} & \multirow{2}[0]{*}{Input} & \multirow{2}[0]{*}{Channel} & \multirow{2}[0]{*}{Stride} & \multicolumn{4}{c}{Scale} \\
    & $n_{\text{min}}$ & $n_{\text{max}}$ &  &  &  &  &  &  & D & W\_I & W\_O & R \\
    \hline
    Conv\_stem & $1$ & $1$ & $3\!\times\!3$ Conv. & N/A & N/A & $224\times224\times3$ & $112\times112\times32$ & $2$ & \XSolidBrush & \XSolidBrush & \Checkmark & \Checkmark \\
    Stage $1$ & $1$ & $1$ & MBConv & $1$ & Y/N & $112\times112\times32$ & $112\times112\times16$ & $1$ & \Checkmark & \Checkmark & \Checkmark & \Checkmark \\
    Stage $2$ & $1$ & $4$ & MBConv & $3/6$ & Y/N & $112\times112\times16$ & $56\times56\times32$ & $2$ & \Checkmark & \Checkmark & \Checkmark & \Checkmark \\
    Stage $3$ & $1$ & $4$ & MBConv & $3/6$ & Y/N & $56\times56\times32$ & $28\times28\times40$ & $2$ & \Checkmark & \Checkmark & \Checkmark & \Checkmark \\
    Stage $4$ & $1$ & $4$ & MBConv & $3/6$ & Y/N & $28\times28\times40$ & $14\times14\times80$ & $2$ & \Checkmark & \Checkmark & \Checkmark & \Checkmark \\
    Stage $5$ & $1$ & $4$ & MBConv & $3/6$ & Y/N & $14\times14\times80$ & $14\times14\times96$ & $1$ & \Checkmark & \Checkmark & \Checkmark & \Checkmark \\
    Stage $6$ & $1$ & $4$ & MBConv & $3/6$ & Y/N & $14\times14\times96$ & $7\times7\times192$ & $2$ & \Checkmark & \Checkmark & \Checkmark & \Checkmark \\
    Stage $7$ & $1$ & $1$ & MBConv & $3/6$ & Y/N & $7\times7\times192$ & $7\times7\times320$ & $1$ & \Checkmark & \Checkmark & \Checkmark & \Checkmark \\
    Conv\_out & $1$ & $1$ & $1\!\times\!1$ Conv. & N/A & N/A & $7\times7\times320$ & $7\times7\times1280$ & $1$ & \XSolidBrush & \Checkmark & \XSolidBrush & \Checkmark \\
    Pooling & $1$ & $1$ & Global avgpool & N/A & N/A & $7\times7\times1280$ & $1280$ & N/A & \XSolidBrush & \XSolidBrush & \XSolidBrush & \XSolidBrush \\
    Classifier & $1$ & $1$ & FC & N/A & N/A & $1280$ & $1000$ & N/A & \XSolidBrush & \XSolidBrush & \XSolidBrush & \XSolidBrush \\
    \hline
    \end{tabular}}
  \label{tab:search_space_base}
\end{table*}

\textbf{Jointly Base Model and Scaling Strategies Searching:} We applied the proposed MCEA for the search with iteration number $T$ as $8$ and sampling number of base models for obtaining initial scaling strategies $S_j^{(0)}$ as $20$. 

We first conducted an initial step for obtaining the $S_j^{(0)}$ based on~($7$) of the main body. Then, we iteratively searched the optimal base model and scaling strategies for $T$ iterations. In each sub-optimization process of the base model search steps, we undertook an evolution algorithm NSGA-II~\cite{deb2002a} with population size $P$ as $50$ and generation size $N$ as $40$. Meanwhile, although we can also employ the evolution algorithm  for scaling strategy search steps, we directly applied a small grid search on the trained super-supernet as the search space of a scaling stage is small enough. Note that we can also use small $N$ with large $T$ (such as $N=8,T=20$), but this may be easy to fall into sub-optimal as the evolution algorithm needs to undertake crossover-mutation steps.

\textbf{Model Retraining:} We followed previous work~\cite{li2020blockwisely,tan2019efficientnet} for obtaining the common training recipe. The models were trained using a RMSProp optimizer with momentum as $0.9$ and weight decay as $1\times10^{-5}$. Initial learning was set as $0.08$ with a step strategy for $300$ epochs. The learning rate was increased from zero to $0.08$ linearly in the first $3$ epochs with batch size $1024$, and then decayed to $0.97$ every $2.4$ epochs. In addition, exponential moving average (EMA) on weights was adopted with a decay rate $0.9999$. RandAugment~\cite{subuk2020randaugment} was also introduced. 

Furthermore, for the ImageNet-$100$ dataset in ablation studies, we modified batch size as $256$, decay rate of the EMA as $0.99$, and epoch number as $200$.

\textbf{Fine-tuning Tasks:} We followed the model retraining settings for ImageNet-$1$k experiments, but removed the EMA and reduced the batch size as $256$.

\begin{table}[!t]
  \centering
  \caption{The range of the search space of the scaling strategies. We take $M=3$ as an example. Depth and width use the same intervals of choices. Scaling stage $0$ refers to the base model.}
    \begin{tabular}{c|c|c|c|c|c|c}
    \hline
    Scaling stage & \multicolumn{3}{c|}{Depth / Width} & \multicolumn{3}{c}{Resolution} \\
    $j$ & Min & Step & Max & Min & Step & Max \\
    \hline
    $0$ & $1.00$ & $0.00$ & $1.00$ & $1.00$ & $0.00$ & $1.00$ \\
    $1$ & $1.04$ & $0.04$ & $1.16$ & $1.00$ & $0.07$ & $1.14$ \\
    $2$ & $1.20$ & $0.04$ & $1.36$ & $1.21$ & $0.07$ & $1.35$ \\
    $3$ & $1.40$ & $0.04$ & $1.64$ & $1.43$ & $0.07$ & $1.57$ \\
    \hline
    \end{tabular}
  \label{tab:search_space_scaling}
\end{table}

\section{Detailed Settings and Definitions of Comparisons of Pearson, Spearman, and Kendall Coefficients} \label{sec:coefficients}

In Table~$4$ of the main body, we evaluated the proposed HSS with three coefficients, including Pearson $\rho_{\text{P}}$~\cite{benesty2009pearson}, Spearman $\rho_{\text{S}}$~\cite{zar2005spearman}, and Kendall $\tau$~\cite{kendall1938a}, compared to the original uniform sampling in~\cite{yu2020bignas}, in order to indicate the super-supernet trained with our HSS can provide more precisely ranking for architectures. In this section, we explain the detailed settings and definitions of them, respectively.

Here, we assume the validation accuracies of the paths sampled the super-supernet should be order preserving on FLOPs same to the actual performance of those trained from scratch. This matches the assumption of the next architecture search step in the one-shot NAS that the validation accuracies of the paths can reflect on the actual performance. Thus, we utilized the three ones to evaluate the effectiveness of our HSS on the order preservation of the path performance. 

We define a validation accuracy vector $\boldsymbol{ACC}=[ACC_1,\cdots,ACC_Q]^{\text{T}}$ and a FLOPs vector $\boldsymbol{f}=[f_1,\cdots,f_Q]^{\text{T}}$ for all the $Q$ sampled paths, where $ACC_q$ and $f_q$ are both for one corresponding path. In our experiments, $Q=6,000$. We can calculate the Pearson coefficient $\rho_{\text{P}}$ by
\begin{align}
    \rho_{\text{P}}=\frac{\text{Cov}(\boldsymbol{ACC}, \boldsymbol{f})}{\sigma_{ACC}\cdot\sigma_{f}}
    =\frac{\text{E}[(\boldsymbol{ACC}-\mu_{ACC})(\boldsymbol{f}-\mu_{f})]}{\sigma_{ACC}\cdot\sigma_{f}},
\end{align}

\noindent where $\text{Cov}(\cdot,\cdot)$ is covariance function, $\sigma_{ACC}$ and $\sigma_{f}$ are the standard deviations of the two vectors, respectively, and $\mu_{ACC}$ and $\mu_{f}$ are the means of the two vectors, respectively.

For the Spearman coefficient $\rho_{\text{S}}$, as it replace the accuracy and FLOPs vectors by their rank vectors $\tilde{\boldsymbol{ACC}}$ and $\tilde{\boldsymbol{f}}$, respectively, we can compute it by
\begin{equation}
    \rho_{\text{S}}=\frac{\text{Cov}(\tilde{\boldsymbol{ACC}}, \tilde{\boldsymbol{f}})}{\sigma_{\tilde{ACC}}\cdot\sigma_{\tilde{f}}},\label{eq:spearman}
\end{equation}

\noindent where $\sigma_{\tilde{ACC}}$ and $\sigma_{\tilde{f}}$ are the standard deviations of the two rank vectors, respectively. As all the ranks are distinct integers,~\eqref{eq:spearman} can be transferred to
\begin{equation}
    \rho_{\text{S}}=1-\frac{6\sum_{q=1}^Q(\tilde{ACC}_q-\tilde{f}_q)^2}{Q(Q^2-1)}.
\end{equation}

Third, the Kendall coefficient $\tau$ is defined as
\begin{equation}
    \tau=\left|\frac{Q_{\text{concordant}}-Q_{\text{disconcordant}}}{Q_{\text{all}}}\right|,
\end{equation}

\noindent where 
\begin{align}
    Q_{\text{concordant}}&=\#\{(ACC_{q_1}-ACC_{q_2})(f_{q_1}-f_{q_2})>0,q_1<q_2\},\\
    Q_{\text{disconcordant}}&=\#\{(ACC_{q_1}-ACC_{q_2})(f_{q_1}-f_{q_2})\le0,q_1<q_2\},
\end{align}

\noindent and $Q_{\text{all}}=\frac{Q(Q-1)}{2}$.

All the three coefficients are the larger the better in the interval of $[0\%,100\%]$. More detailed explanations of them can be found in Wiki Pedia for Pearson\footnote{\url{https://en.wikipedia.org/wiki/Pearson_correlation_coefficient}.}, Spearman\footnote{\url{https://en.wikipedia.org/wiki/Spearman\%27s_rank_correlation_coefficient}.}, and Kendall\footnote{\url{https://en.wikipedia.org/wiki/Kendall_rank_correlation_coefficient}.} coefficients.

\section{Definitions of Compared Functions in Lager-scale Architecture Generalization} \label{sec:definition_functions_generalization}

We have compared the proposed larger-scale architecture generalization function in the ablation studies with linear and squared functions. Here, we define the forms of the two compared functions and their optimization processes in the ablation studies.

We define the optimal scales of $M+1$ scaling stages as $\{\boldsymbol{\hat{S}}_j\}_{j=0}^M$. We also pre-define $\boldsymbol{\hat{S}}_0$ as $\hat{d}_0=\hat{w}_0=\hat{r}_0=1$ for the base model. These are same to those in the main body.

\textbf{Linear Function}. Here, we define the linear functions for the three dimensions, respectively, as
\begin{equation}
    \left\{
        \begin{array}{l}
            \hat{d}_j=a_0^{(d)}\cdot j+a_1^{(d)}\\
            \hat{w}_j=a_0^{(w)}\cdot j+a_1^{(w)}\\
            \hat{r}_j=a_0^{(r)}\cdot j+a_1^{(r)}
        \end{array}
    \right.,\label{eq:dwr_linear}
\end{equation}

\noindent where $a_0$ and $a_1$ are parameters.

In order to guarantee the relation between $j=0$ and $\boldsymbol{\hat{S}}_0$, we put $\hat{d}=\hat{w}=\hat{r}=1,j=0$ into~\eqref{eq:dwr_linear} and obtain
\begin{equation}
    \left\{
        \begin{array}{l}
            a_1^{(d)}=1\\
            a_1^{(w)}=1\\
            a_1^{(r)}=1
        \end{array}
    \right..\label{eq:dwr_linear_a1}
\end{equation}

After that, we re-put~\eqref{eq:dwr_linear_a1} into~\eqref{eq:dwr_linear} and obtain the linear functions for the three dimensions as
\begin{equation}
    \left\{
        \begin{array}{l}
            \hat{d}_j=a_0^{(d)}\cdot j+1\\
            \hat{w}_j=a_0^{(w)}\cdot j+1\\
            \hat{r}_j=a_0^{(r)}\cdot j+1
        \end{array}
    \right..\label{eq:dwr_linear_final}
\end{equation}

We should note that $a_0^{(d)}$, $a_0^{(w)}$, and $a_0^{(r)}$ are all constrained to be positive as the trends of the three dimensions should be monotonically increasing.

\textbf{Squared Function}. Similarly, we define the squared functions for the three dimensions, respectively, as
\begin{equation}
    \left\{
        \begin{array}{l}
            \hat{d}_j=a_0^{(d)}\cdot j^2+a_1^{(d)}\cdot j+a_2^{(d)}\\
            \hat{w}_j=a_0^{(w)}\cdot j^2+a_1^{(w)}\cdot j+a_2^{(w)}\\
            \hat{r}_j=a_0^{(r)}\cdot j^2+a_1^{(r)}\cdot j+a_2^{(r)}
        \end{array}
    \right.,\label{eq:dwr_squared}
\end{equation}

\noindent where $a_0$, $a_1$ ,and $a_2$ are parameters.

In the same way, we put $\hat{d}=\hat{w}=\hat{r}=1,j=0$ into~\eqref{eq:dwr_squared} and obtain
\begin{equation}
    \left\{
        \begin{array}{l}
            a_2^{(d)}=1\\
            a_2^{(w)}=1\\
            a_2^{(r)}=1
        \end{array}
    \right..\label{eq:dwr_squared_a2}
\end{equation}

We also re-put~\eqref{eq:dwr_squared_a2} into~\eqref{eq:dwr_squared} and obtain the squared functions as
\begin{equation}
    \left\{
        \begin{array}{l}
            \hat{d}_j=a_0^{(d)}\cdot j^2+a_1^{(d)}\cdot j+1\\
            \hat{w}_j=a_0^{(w)}\cdot j^2+a_1^{(w)}\cdot j+1\\
            \hat{r}_j=a_0^{(r)}\cdot j^2+a_1^{(r)}\cdot j+1
        \end{array}
    \right.,\label{eq:dwr_squared_final}
\end{equation}

\noindent where $a_0^{(d)}$, $a_0^{(w)}$, and $a_0^{(r)}$ are also constrained to be positive.

\textbf{Optimization Process}. We define the optimization objective for one dimension with a function $\text{func}(\cdot;a_0,a_1)$ as
\begin{align}
    \argmin_{a_0,a_1}&\sum_{j=1}^{M}||y_j-\text{func}(j;a_0,a_1)||^2,\\
    \text{s.t.}&\ a_0>0,\nonumber
\end{align}

\noindent where $\text{func}(\cdot;a_0,a_1)$ is selected from~\eqref{eq:dwr_linear_final},~\eqref{eq:dwr_squared_final}, or (8) of the main body and $y_j$ is the searched value of one dimension ($\hat{d}$, $\hat{w}$, or $\hat{r}$) in scaling stage $j$. We can directly optimize by stochastic gradient descent (SGD) or other optimization algorithms.

\section{Search Space of Base Model}\label{ssec:base_model_search_space}

We introduce the search space of base model for model scaling in Table~\ref{tab:search_space_base}. Super-supernet training will sample base models in paths according to the search space. Following previous work~\cite{su2021prioritized,you2020greedynas}, we utilize mobile inverted bottleneck MBConv~\cite{sandler2018mobilenetv2} with expand rates as $3$ or $6$. Squeeze-and-excitation network (SENet)~\cite{hu2018squeeze} is also applied in the search space. No more than four layers are assigned in each stage for the base model. Specifically, at least one layer should be selected in each path of super-supernet training. 

\section{Search Space of Scaling Strategies}\label{ssec:scaling_strategy_search_space}

The search space of scaling strategy is defined in Table~\ref{tab:search_space_scaling}. We take maximum scaling stage number $M=3$ as an example. Note that scaling stage $j=0$ represents the base model. As shown in Table~\ref{tab:search_space_scaling}, the three dimensions,~\emph{i.e.}, depth, width, and resolution, are set as one for the base model, respectively. For larger scaling stages, depth and width are selected in the same intervals, while resolution has other different ones, which are empirically designed. 

For the three dimensions, we first set the center points of the intervals by Monte Carlo simulation according to the FLOPs budgets, and then obtain the ranges of the intervals based on the center points, respectively. In a dimension, we define same value of steps for each scaling stage as $0.04$ for both depth and width, and $0.07$ for resolution. For the former ones, a certain and suitable step is enough for the super-supernet training, while for the latter one, we empirically set them according to BigNAS~\cite{yu2020bignas}. Especially for width, we need to guarantee the channel numbers of different paths to uniformly change as we used the technique in~\cite{su2021bcnet} for width in super-supernet training. 

According to the search space of base model in Table~\ref{tab:search_space_base} and that of scaling strategy in Table~\ref{tab:search_space_scaling}, we can visualize the the whole search space to FLOPs by Monte Carlo simulation, as shown in Figure~$3$ of the main body for ImageNet-$1$k experiments and Figure~\ref{fig:search_space_ablation} for ablation studies on ImageNet-$100$ (half channel numbers for each layer). Our search spaces are both in similar multi-modal forms of the histograms, instead of a simple bell-shape form of the original uniform sampling. Note that although the respective sampling distributions of each scaling stage are overlapped with contiguous ones, the paths near the boundaries between scaling stages (\emph{i.e.}, away from the FLOPs Budgets) will be dropped in the search, as they do not satisfy the FLOPs budgets. This means that they do not affect the joint base model and scaling strategy search after the super-supernet training.

\section{Path Generation with Sampled Base Model and Scaling Strategy}

In the super-supernet training, we specifically sample a base model architecture by uniform sampling and a scaling strategy by the proposed hierarchical sampling strategy (HSS) for each batch. Thus, we should combine them and generate a path for the super-supernet training.

For the depth, we simply copy the structure of the last non-identity block in each stage that can be scaled (See line ``D'' in Table~\ref{tab:search_space_base}). ``Conv\_stem'', ``Pooling'', ``Conv\_out'', and ``Classifier'' cannot be scaled in depth. The number $n_{\text{s}}$ of scaled blocks is defined as
\begin{equation}
    n_{\text{s}}=\left\lceil n\times d\right\rceil,
\end{equation}
where $n$ is the block number of a stage in a sampled path and $d$ is depth coefficient.

Similarly for the width, we directly combine the added channels into each layer that can be scaled in the base model. Note that the input channel number of ``Conv\_stem'' (three channels for RGB) and the output channel number of ``Conv\_out'' ($1280$ channels same to the input channel number of the fully connected layer) are not modified (See line ``W\_I'' and ``W\_O'' in Table~\ref{tab:search_space_base}). The latter one is different with that of EfficientNet~\cite{tan2019efficientnet}. The scaled channel number $c_{\text{s}}$ of a layer is defined as
\begin{equation}
    c_{\text{s}}=\left\lceil c\times w\right\rceil,
\end{equation}
where $c$ is the channel number of a layer in a sampled path and $w$ is width coefficient.

For the resolution, we directly resize the input images before augmentations to the target size. It influences all the layers before ``Pooling''. The scaled resolution $r_{\text{s}}$ of a input image is defined as
\begin{equation}
    r_{\text{s}}=\left\lceil 224\times r\right\rceil,
\end{equation}
where $r$ is resolution coefficient.

The size of total search space in the super-supernet training is almost $2\times10^{26}$.

\section{Search Cost Computation of Referred Methods}\label{sec:search_cost_referred}

\textbf{EfficientNet}~\cite{tan2019efficientnet}: As they said in the paper that they conducted a small grid search for $d_1$, $w_1$, and $r_1$ based on the constraint $d_1\times w_1^2\times r_1^2\approx2$, we sampled them in the interval of $[1,2]$ with step as $0.01$ and analyzed all the cases. The ones that satisfy the constraint $|d_1\times w_1^2\times r_1^2-2|\le0.1$ were counted and $10,285$ cases were included. For the training time of a model, we found that training an EfficientNet-B$4$ model ($4.2$G FLOPs) cost about one TPU day. Thus, training an EfficientNet-B$1$ model ($0.7$G FLOPs) may cost about $1/6$ TPU days. The total search cost for the compound scaling should be $1,714$ TPU days.

Another work~\cite{wan2020fbnetv2} said the lower bound of searching EfficientNet-B$0$ is $91,000$ TPU hours,~\emph{i.e.}, $3,792$ TPU days. However, we did not consider the part.

\textbf{EfficientNet-X}~\cite{li2021searching}: They undertook a two-level grid search for $d_1$, $w_1$, and $r_1$. We first sampled them in the interval of $[1,2]$ with step as $0.1$ and remained all the cases. Then, a smaller grid search around the best candidate with difference ranges in the interval of $[-0.1,0.1]$ and step $0.01$. Finally, we can obtain $11^3+21^3-1=10,591$ cases. Similar to the EfficientNet, the total search cost should be $1,756$ TPU days.

\textbf{BigNAS}~\cite{yu2020bignas}: As discussed in the paper, they train the supernet for $36$ hours with $64$ TPUv$3$. Here, the model sized from $200$M to $2,000$M FLOPs and they searched from $200$M to $1,000$M FLOPs. Thus, for searching a $10$G FLOPs architecture, the supernet should size from $200$M to $20$G FLOPs,~\emph{i.e.}, $100\times$ than before. The supernet training time should be also enlarged to $10\times$ as $36\times64\times10=23,040$ TPU hours,~\emph{i.e.}, $960$ TPU days. Although the paper did not mention their cost on searching architectures, we assumed their cost ratio of supernet training and searching is same to ours and approximated their searching cost as $960\times(106/379)=268$ TPU days.

\textbf{FBNetV$2$}~\cite{wan2020fbnetv2}: As shown in the paper, the authors searched the FBNetV$2$-L$1$ model ($0.33$M FLOPs) with total $0.6$k GPU hours. We linearly expand the number by FLOPs and obtain $39,182$ GPU hours,~\emph{i.e.}, $1,633$ GPU days.

\textbf{OFA}~\cite{cai2020onceforall}: They first searched a large architecture ($40$ GPU hours), where FLOPs budget was not mentioned in the paper and we assumed as $600$M FLOPs, then trained the model for enough time ($1,200$ GPU hours), and searched target ones from the large one by progressive shrinking and fine-tuning ($75\times40$ GPU hours). For searching and training a $10$G FLOPs large model, they need $(1,200+40)\times10,000/600=20,667$ GPU hours. Then, searching the five scaling stages needs at least $75\times40\times5=15,000$ GPU hours. The total time cost should be $20,667+15,000=35,667$ GPU hours,~\emph{i.e.}, $1,486$ GPU days.

\textbf{MnasNet}~\cite{tan2019mnasnet}: Their total search cost is $91,000$ TPU hours for $388$M FLOPs budget. Transferring to our total FLOPs budgets as $10,000+6,000+3,000+1,500+800+350=21,650$M FLOPs, their search cost should be  $91,000\times21,650/388=5,077,706$ TPU hours,~\emph{i.e.}, $211,571$ TPU days.

\textbf{Speed comparisons of Different Devices}: We have reviewed the computation speed of different devices, including V$100$ GPU and TPUv$2$/$3$, which were used in previous work~\cite{cai2020onceforall,li2021searching,tan2019mnasnet,tan2019efficientnet,wan2020fbnetv2,yu2020bignas} or this paper. TPUv$3$ is $2.7\times$ faster than TPUv$2$\footnote{See \url{https://www.linleygroup.com/newsletters/newsletter_detail.php?num=6203&year=2020&tag=3}.}. The speed of a TPUv$2$ core is similar to that of one V$100$\footnote{See page $24$ in \url{https://storage.googleapis.com/nexttpu/index.html}.}. Combining all of them, we can obtain the training speed ratios as $\text{a TPUv}3\text{ core}:\text{a TPUv}2\text{ core}:\text{a V}100=2.7:1:1$.

Note that we only discuss the search cost, as we assume all the methods require same cost on retraining, except for the BigNAS, which does not need to retrain the searched architectures. Retraining needs about $280$ GPU days on V$100$ and the ScaleNet totally costs $765$ GPU days, which is much smaller than $1,228$ TPU days on TPUv$3$ of the BigNAS.

\begin{table}[!t]
  \centering
  \caption{The searched and generalized scaling strategies from ScaleNet-S$0$ to ScaleNet-S$5$.}
    \begin{tabular}{c|c|c|c}
    \hline
    Scaling stage & Depth & Width & Resolution \\
    \hline
    $0$ & $1.000$ & $1.000$ & $1.000$ \\
    $1$ & $1.080$ & $1.040$ & $1.140$ \\
    $2$ & $1.360$ & $1.200$ & $1.355$ \\
    $3$ & $1.480$ & $1.400$ & $1.580$ \\
    $4$ & $1.653$ & $1.534$ & $2.042$ \\
    $5$ & $1.848$ & $1.688$ & $2.378$ \\
    \hline
    \end{tabular}
  \label{tab:scaling_strategy}
\end{table}

\begin{figure*}[!t]
    \begin{center}
        \includegraphics[width=.85\linewidth]{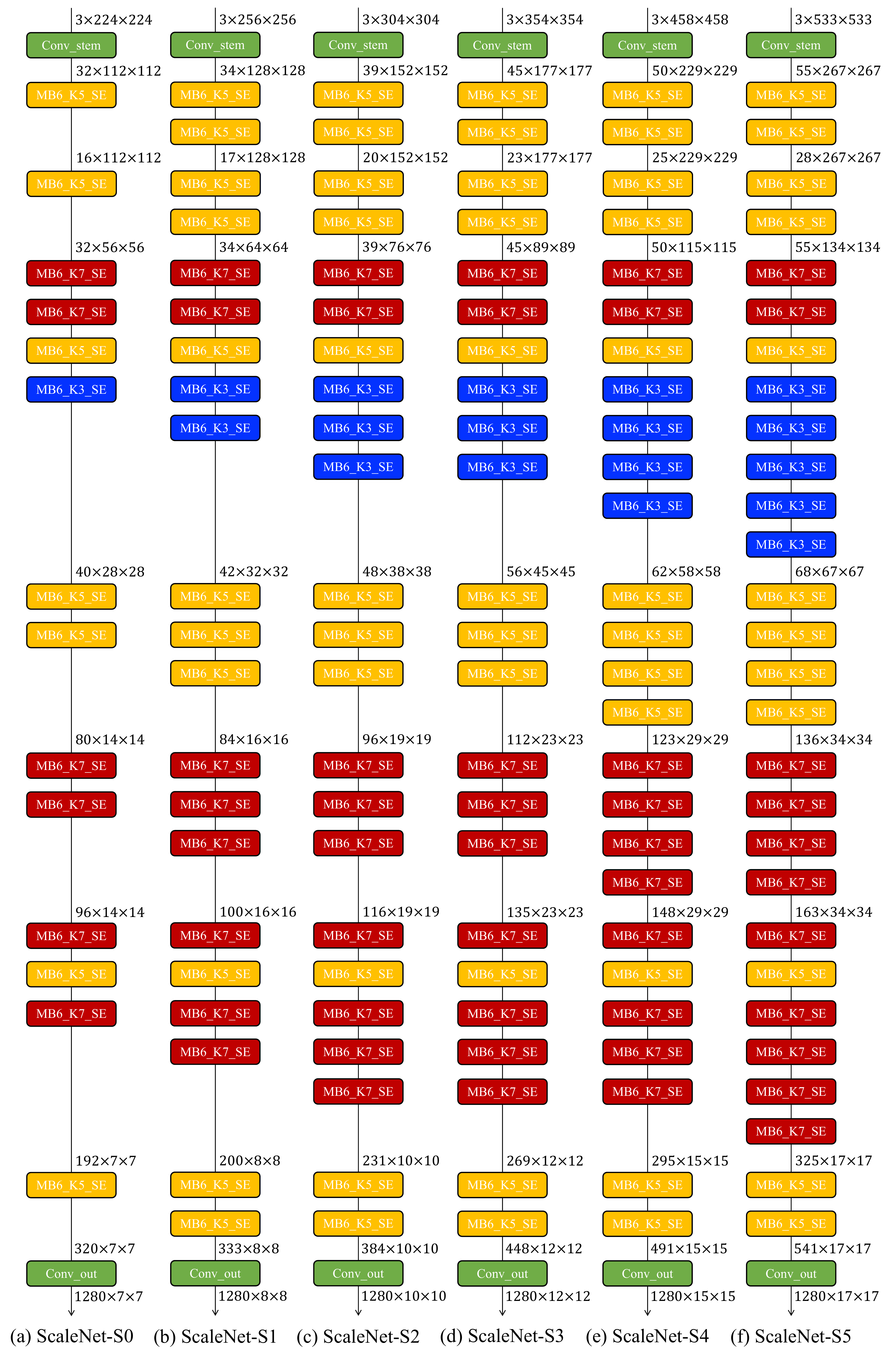}
    \end{center}
    \caption{Visualizations of searched and generalized architectures from ScaleNet-S$0$ to ScaleNet-S$5$. ``MB$e$\_K$k$\_SE'' means the block uses $k\times k$ depth-wise convolution as the intermediate convolution with expand rate as $e$ and SE block. The architectures are split into stages (not scaling stages) by the resolutions. All the downsampling blocks are employed in the first layer of each stage.}
    \label{fig:visualization}
\end{figure*}

\section{Visualizations of Search Architectures}\label{sec:visualization}

Table~\ref{tab:scaling_strategy} shows the searched and generalized scaling strategies of the proposed ScaleNet for the ImageNet-$1$k experiments. Moreover, Figure~\ref{fig:visualization} shows the corresponding searched and generalized architectures. Here, we find that they have the following three properties:

\begin{itemize}
    \item Regarding kernels of convolutional layers, the base model usually prefers to use more large kernels ($5\times5$ or $7\times7$ convolution layers) and more $7\times7$ kernels are applied in the deeper layers. 
    
    \item With different FLOPs budgets, two architectures may have same numbers of layers, but with distinct width and resolution, respectively. As different values of depth can respond to a same architecture.
    
    \item Regarding resolutions of various scaling stages, the ScaleNet tends to use higher resolutions than those in~\cite{dollar2021fast,tan2019efficientnet} to improve the performance. They can also reduce the numbers of parameters under certain FLOPs budgets to some extent.
\end{itemize}

\begin{table*}[!t]
  \centering
  \caption{Detailed comparisons between our ScaleNet and EfficientNet. ``re-impl'' means we reimplemented the model with our settings. ``$*$'' used our scaling strategies with EfficientNet-B$0$. The best results are highlighted in \textbf{bold}.}
    \resizebox{\linewidth}{!}{
    \begin{tabular}{lccccccc}
    \hline
    \multicolumn{1}{c}{Model} & Top-$1$ acc. (\%) & Top-$5$ acc. (\%) & \#Params (M) & FLOPs (G) & Resolution \\
    \hline
    EfficientNet-B$1$~\cite{tan2019efficientnet} & $78.8$ & $94.4$ & $7.8$ & $0.70$ & $240\!\times\!240$ \\
    EfficientNet-B$1$ (re-impl)                 & $78.7$ & $94.1$ & $7.8$ & $0.70$ & $240\!\times\!240$ \\
    EfficientNet-S$1^*$                          & $79.2$ & $94.6$ & $8.3$ & $0.79$ & $256\!\times\!256$ \\
    \bf ScaleNet-S$\boldsymbol{1}$               & $\boldsymbol{79.9}$ & $\boldsymbol{94.8}$ & $\boldsymbol{7.4}$ & $\boldsymbol{0.80}$ & $\boldsymbol{256\!\times\!256}$ \\
    \hline
    EfficientNet-B$2$~\cite{tan2019efficientnet} & $79.8$ & $94.9$ & $9.2$ & $1.00$ & $260\!\times\!260$ \\
    EfficientNet-B$2$ (re-impl)                 & $80.4$ & $95.1$ & $9.2$ & $1.00$ & $260\!\times\!260$ \\
    EfficientNet-S$2^*$                          & $80.8$ & $95.4$ & $11.8$ & $1.58$ & $304\!\times\!304$ \\
    \bf ScaleNet-S$\boldsymbol{2}$               & $\boldsymbol{81.3}$ & $\boldsymbol{95.6}$ & $\boldsymbol{10.2}$ & $\boldsymbol{1.45}$ & $\boldsymbol{304\!\times\!304}$ \\
    \hline
    EfficientNet-B$3$~\cite{tan2019efficientnet} & $81.1$ & $95.5$ & $12.0$ & $1.80$ & $300\!\times\!300$ \\
    EfficientNet-B$3$ (re-impl)                 & $81.1$ & $95.4$ & $12.0$ & $1.80$ & $300\!\times\!300$ \\
    EfficientNet-S$3^*$                          & $82.1$ & $95.8$ & $15.4$ & $2.94$ & $354\!\times\!354$ \\
    \bf ScaleNet-S$\boldsymbol{3}$               & $\boldsymbol{82.2}$ & $\boldsymbol{95.9}$ & $\boldsymbol{13.2}$ & $\boldsymbol{2.76}$ & $\boldsymbol{354\!\times\!354}$ \\
    \hline
    EfficientNet-B$4$~\cite{tan2019efficientnet} & $82.6$ & $96.3$ & $19.0$ & $4.20$ & $380\!\times\!380$ \\
    EfficientNet-B$4$ (re-impl)                 & $82.6$ & $96.3$ & $19.0$ & $4.20$ & $380\!\times\!380$ \\
    EfficientNet-S$4^*$                          & $82.8$ & $96.3$ & $19.5$ & $6.34$ & $458\!\times\!458$ \\
    \bf ScaleNet-S$\boldsymbol{4}$               & $\boldsymbol{83.2}$ & $\boldsymbol{96.6}$ & $\boldsymbol{16.1}$ & $\boldsymbol{5.97}$ & $\boldsymbol{458\!\times\!458}$ \\
    \hline
    EfficientNet-B$5$~\cite{tan2019efficientnet} & $83.3$ & $96.7$ & $30.0$ & $9.90$ & $456\!\times\!456$ \\
    EfficientNet-B$5$ (re-impl)                 & $83.2$ & $96.7$ & $30.0$ & $9.90$ & $456\!\times\!456$ \\
    EfficientNet-S$5^*$                          & $83.2$ & $96.4$ & $25.8$ & $11.42$ & $533\!\times\!533$ \\
    \bf ScaleNet-S$\boldsymbol{5}$               & $\boldsymbol{83.7}$ & $\boldsymbol{97.1}$ & $\boldsymbol{20.9}$ & $\boldsymbol{10.22}$ & $\boldsymbol{533\!\times\!533}$ \\
    \hline
    \end{tabular}}
  \label{tab:params_flops}
\end{table*}
\setlength{\tabcolsep}{1.4pt}

\section{Discussion of Maximum Scaling Stage in Searching}

In this section, we discuss the maximum scaling stage issue in the joint search of base model and scaling strategy. As discussed in the ablation studies, searching on more scaling stages can obtain better scaling strategies. This is given that directly searching the scaling strategy on a scaling stage should have better performance than generalizing it. Although we applied the maximum scaling stage $M=3$ and generalized the model on scaling stage $4$ in the ablation studies, using $M=4$ and searching the fourth scaling stage may obtain a better one. 

However, larger $M$ implies consuming more resources, for example, training the super-supernet under current settings on ImageNet-$1$k needs $32$ V$100$ cards with automatic mixed precision (AMP) or $64$ V$100$ cards with full precision. $M$ influences the computational complexity in both super-supernet training and searching. It is approximately exponential to the computational complexity (the computational complexity is proportional to FLOPs). However, with its increase, performance improvement of the generalized models should be approximately logarithmically increased as in Table $3$. As current experimental results have reached state-of-the-art performance, utilizing double resources to training the super-supernet and searching with one more scaling stage and achieving $0.5\%$ top-$1$ accuracy improvement on ImageNet-$1$k (similar to s$4$ models in Table $3$ of the main body. We also trained an s$5$ model and found that its top-$1$ accuracy is $90.74\%$, slightly higher than $90.46\%$ of s$4$ model.) between a directly searched one and a generalized one may not be necessary. Since the performance of models tends to saturate with the increase of model scales, in practice a moderate $M$ should be favored for better and efficient search. It can be theoretically set as infinite, but a too large value of it is worthless. Thus, we only set $M$ as three in the experiments.

\begin{figure*}[!t]
    \centering
    \begin{subfigure}[t]{.49\linewidth}
        \begin{center}
            \includegraphics[width=\linewidth]{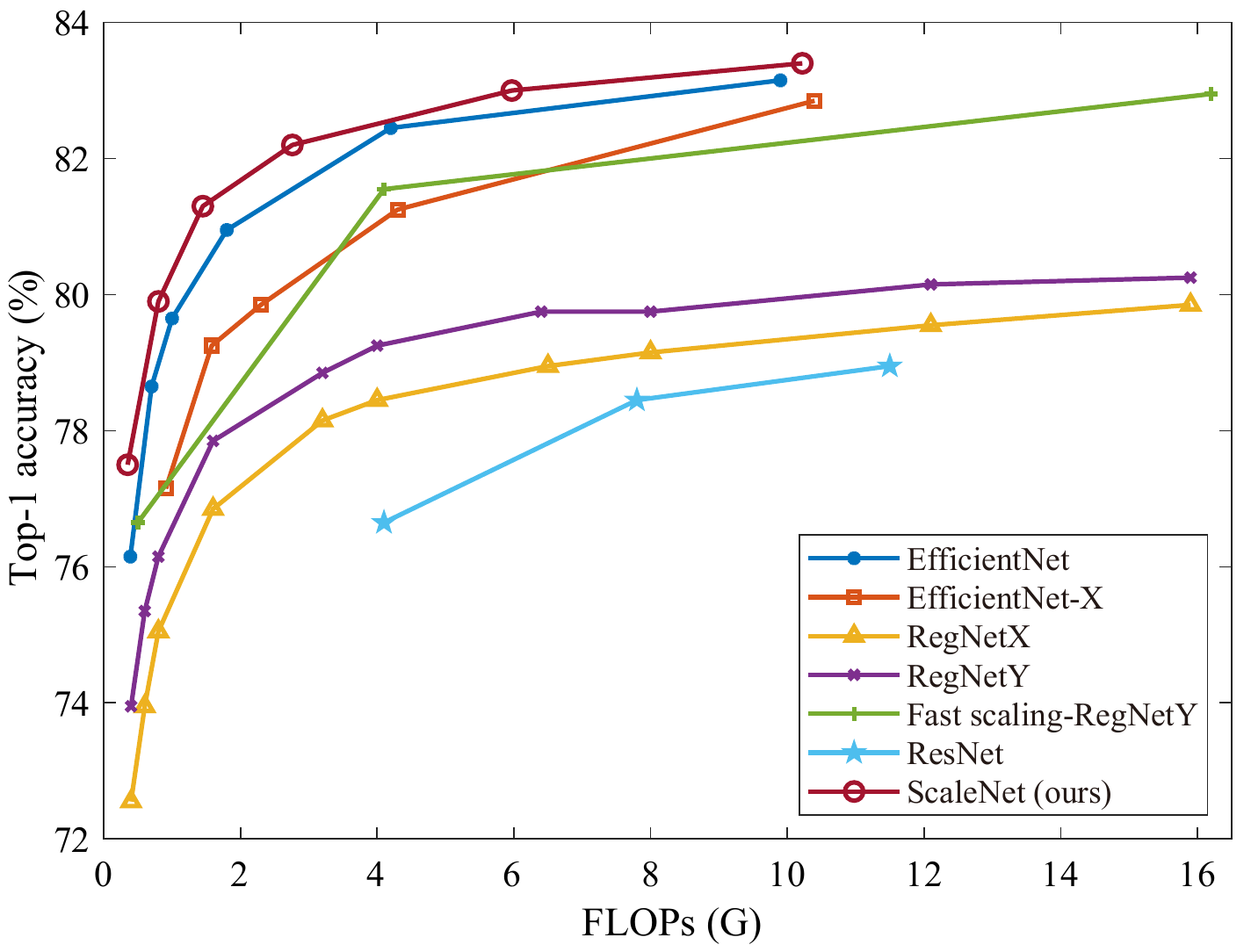}
        \end{center}
        \subcaption{FLOPs versus top-$1$ accuracy.}
    \end{subfigure}
    \begin{subfigure}[t]{.49\linewidth}
        \begin{center}
            \includegraphics[width=\linewidth]{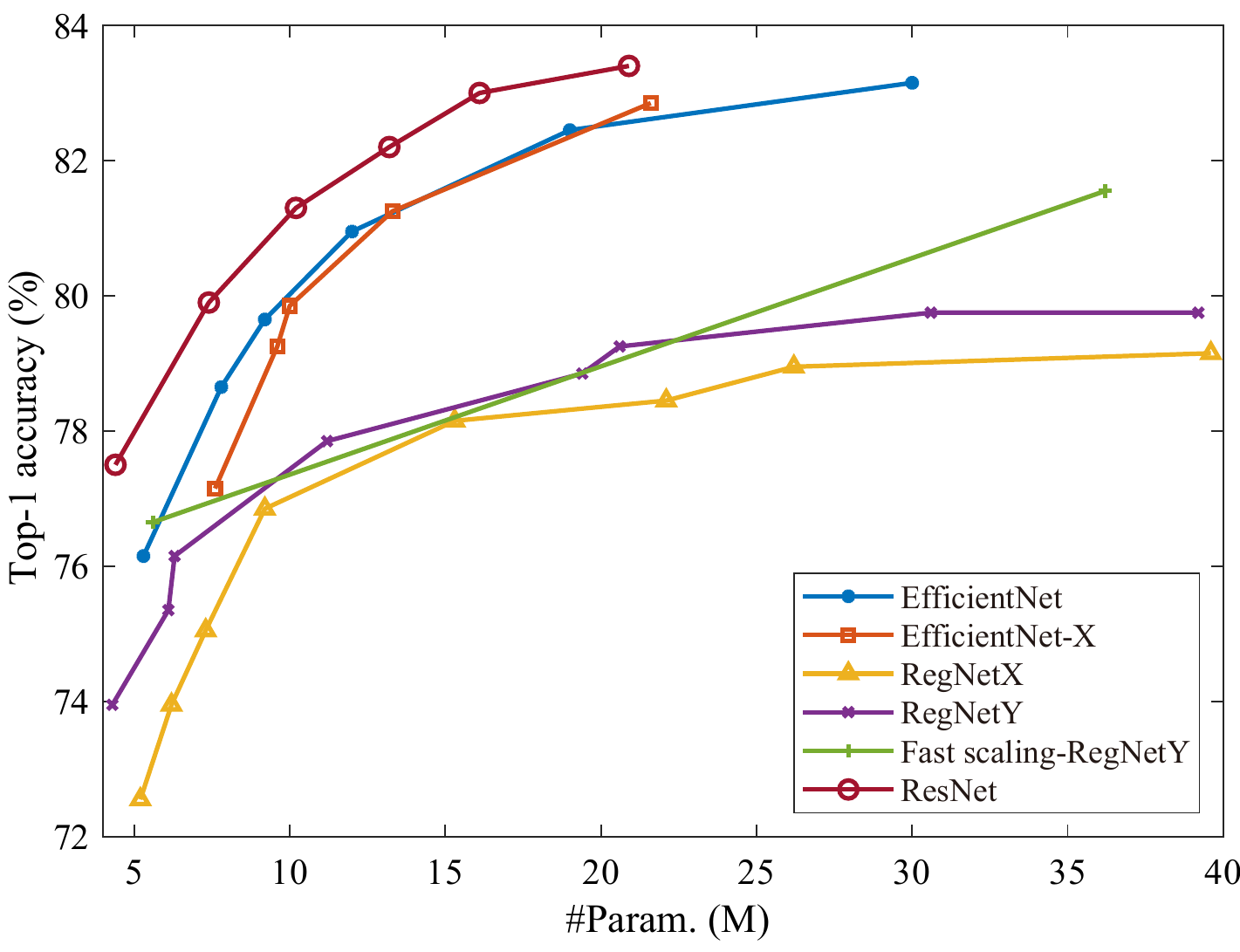}
        \end{center}
        \subcaption{\#Param. versus top-$1$ accuracy.}
    \end{subfigure}
    \caption{Comparisons between our ScaleNet and other baseline methods with on ImageNet-$1$k dataset. Both (a) FLOPs and (b) number of parameters (\#Param.) are compared versus top-$1$ accuracy. Note that we do not illustrate ResNet in (b) as it has too many parameters.}
    \label{fig:flops_acc_i1k}
\end{figure*}

\section{Comparisons with EfficientNet}

Table~\ref{tab:params_flops} shows detailed comparisons between our ScaleNet and EfficientNet~\cite{tan2019efficientnet}. With same experimental settings, the searched and generalized models of our ScaleNet can significantly outperform the re-implemented EfficientNet models in each scaling stage with similar FLOPs and numbers of parameters, respectively. As our models were implemented with larger resolutions than EfficientNet ones, we can slightly reduce the numbers of parameters for each.

Meanwhile, we conducted a group of experiments on our scaling strategies with EfficientNet-B$0$ as base model, namely EfficientNet-S$1$ to EfficientNet-S$5$. Compared with the baseline models, they can achieve remarkable or marginal performance improvement in various scaling stages. This means our obtained scaling strategies is better than those of the baseline. Furthermore, our results can also surpass those of them, which demonstrates our searched base model is more suitable for scaling than the EfficientNet-B$0$.

To intuitively compare the performance of our ScaleNet with other baseline methods, we visualize the top-$1$ accuracies of our ScaleNet~\emph{w.r.t.}~FLOPs and number of parameters, respectively, in Figure~\ref{fig:flops_acc_i1k}. As shown in Figure~\ref{fig:flops_acc_i1k}(a), the performance of the searched and generalized models can outperform all the others as an upper bound. Similarly in Figure~\ref{fig:flops_acc_i1k}(b), our models also perform best among the others with a larger margin on accuracy.

\begin{table}[!t]
  \centering
  \caption{Statistics of five fine-tuning datasets. ``\#Training'', ``\#Test'', and ``\#Class'' are numbers of training samples, test samples, and classes, respectively.}
    \begin{tabular}{c|ccc}
    \hline
    Dataset & \#Training & \#Test & \#Class \\
    \hline
    FGVC Aircraft~\cite{maji2013finegrained} & $6,667$ & $3,333$ & $100$ \\
    Stanford Cars~\cite{krause20133d} & $8,144$ & $8,041$ & $196$ \\
    Food-$101$~\cite{bossard2014food101} & $75,750$ & $25,250$ & $101$ \\
    CIFAR-$10$~\cite{krizhevsky09cifar} & $50,000$ & $10,000$ & $10$ \\
    CIFAR-$100$~\cite{krizhevsky09cifar} & $50,000$ & $10,000$ & $100$ \\
    \hline
    \end{tabular}
  \label{tab:downstream_datasets}
\end{table}

\setlength{\tabcolsep}{14pt}
\begin{table}[!t]
  \centering
  \caption{Comparisons with other state-of-the-art methods in five fine-tuning tasks. Top-$1$ accuracies (\%), FLOPs (G), numbers of parameters (\#Param., M) are reported. The best results on each dataset are highlighted in \textbf{bold}.}
    \begin{tabular}{l@{}l@{}c@{}c@{}c}
    \hline
    \multicolumn{1}{c}{Dataset} & \multicolumn{1}{c}{Model} &\ \ Top-$1$\ \ &\ \ FLOPs\ \ &\ \ \#Param.\ \ \\
    \hline
    \multirow{6}[0]{*}{FGVC Aircraft\ \ } 
    & EfficientNet-B$3$~\cite{tan2019efficientnet} & $90.7$ & $1.80$ & $10$ \\
    & NAT-M$4$~\cite{lu2021neural} & $90.8$ & $0.58$ & $5$ \\
    & Inception-v$4$~\cite{szegedy2017inceptionv4} & $90.9$ & $13.00$ & $41$ \\
    & ScaleNet-S$3$ (ours) & $91.4$ & $2.76$ & $11$ \\
    & ScaleNet-S$4$ (ours) & $92.8$ & $5.97$ & $14$ \\
    & \bf ScaleNet-S$\boldsymbol{5}$ (ours) & $\boldsymbol{92.9}$ & $\boldsymbol{10.22}$ & $\boldsymbol{19}$ \\
    \hline
    \multirow{9}[0]{*}{Stanford Cars} 
    & NAT-M$4$~\cite{lu2021neural} & $92.9$ & $0.37$ & $6$ \\
    & Inception-v$4$~\cite{szegedy2017inceptionv4} & $93.4$ & $13.00$ & $41$ \\
    & EfficientNet-B$3$~\cite{tan2019efficientnet} & $93.6$ & $1.80$ & $10$ \\
    & EfficientNetV2-S~\cite{tan2021efficientnetv2} & $93.8$ & $8.80$ & $24$ \\
    & EfficientNet-B$7$~\cite{tan2019efficientnet} & $94.7$ & $37.00$ & $64$ \\
    & DAT~\cite{ngiam2018domain} & $94.8$ & - & - \\
    & ScaleNet-S$3$ (ours) & $94.4$ & $2.76$ & $11$ \\
    & ScaleNet-S$4$ (ours) & $95.0$ & $5.97$ & $14$ \\
    & \bf ScaleNet-S$\boldsymbol{5}$ (ours) & $\boldsymbol{95.1}$ & $\boldsymbol{10.22}$ & $\boldsymbol{19}$ \\
    \hline
    \multirow{6}[0]{*}{Food-$101$} 
    & NAT-M$4$~\cite{lu2021neural} & $89.4$ & $0.36$ & $5$ \\
    & Inception-v$4$~\cite{szegedy2017inceptionv4} & $90.8$ & $13.00$ & $41$ \\
    & EfficientNet-B$4$~\cite{tan2019efficientnet} & $91.5$ & $4.20$ & $17$ \\
    & ScaleNet-S$3$ (ours) & $91.2$ & $2.76$ & $11$ \\
    & ScaleNet-S$4$ (ours) & $92.0$ & $5.97$ & $14$ \\
    & \bf ScaleNet-S$\boldsymbol{5}$ (ours) & $\boldsymbol{92.2}$ & $\boldsymbol{10.22}$ & $\boldsymbol{19}$ \\
    \hline
    \multirow{5}[0]{*}{CIFAR-$10$} 
    & Proxyless-G+c/o~\cite{cai2018proxylessnas} & $97.2$ & - & $6$ \\
    & NASNet-A~\cite{zoph2018learning} & $98.0$ & $42.00$ & $85$ \\
    & EfficientNet-B$0$~\cite{tan2019efficientnet} & $98.1$ & $0.39$ & $4$ \\
    & NAT-M$3$~\cite{lu2021neural} & $98.2$ & $0.39$ & $6$ \\
    & \bf ScaleNet-S$\boldsymbol{0}$ (ours) & $\boldsymbol{98.3}$ & $\boldsymbol{0.35}$ & $\boldsymbol{3}$ \\
    \hline
    \multirow{4}[0]{*}{CIFAR-$100$} 
    & NASNet-A~\cite{zoph2018learning} & $87.5$ & $42.00$ & $85$ \\
    & NAT-M$3$~\cite{lu2021neural} & $87.7$ & $0.49$ & $8$ \\
    & EfficientNet-B$0$~\cite{tan2019efficientnet} & $88.1$ & $0.39$ & $4$ \\
    & \bf ScaleNet-S$\boldsymbol{0}$ (ours) & $\boldsymbol{88.4}$ & $\boldsymbol{0.35}$ & $\boldsymbol{3}$ \\
    \hline
    \end{tabular}
  \label{tab:downstream2}
\end{table}
\setlength{\tabcolsep}{1.4pt}

\section{More Comparisons on Fine-tuning Tasks}

The statistics of five fine-tuning datasets are shown in Table~\ref{tab:downstream_datasets}, including numbers of training samples, test samples, and classes. Table~\ref{tab:downstream2} shows experimental results on the datasets. On FGVC Aircraft, Stanford Cars, and Food-$101$, we trained ScaleNet-S$3/4/5$ with our ImageNet-pretrained models and obtained state-of-the-art performance. Furthermore, on the other two CIFAR datasets, we can also achieve superior accuracies compared with the baseline models.



    
    
    

\section{Definition Difference between ``Stage'' and ``Scaling Stage''}

``Stage'' and ``scaling stage'' are in two different dimensions. ($1$) ``Stage'' corresponds to depth in one model. As shown in Figure~\ref{fig:visualization}, a model can be divided into seven stages with different channel numbers. ($2$) ``Scaling stage'' is a holistic concept that refers to various models with different FLOPs budgets,~\emph{i.e.}, the subfigures in Figure~\ref{fig:visualization}.

\clearpage
%
%
\bibliographystyle{splncs04}
\bibliography{scalenet-arxiv}
\end{document}